\documentclass{article}
\usepackage{iclr2023_conference,times}

\usepackage{amsmath,amsfonts,bm}

\def\eqref#1{equation~\ref{#1}}

\def\1{\bm{1}}

\DeclareMathAlphabet{\mathsfit}{\encodingdefault}{\sfdefault}{m}{sl}
\SetMathAlphabet{\mathsfit}{bold}{\encodingdefault}{\sfdefault}{bx}{n}

\usepackage{makecell}

\usepackage{hyperref}
\usepackage{url}
\usepackage{microtype}
\usepackage{graphicx}
\usepackage{booktabs}
\usepackage{balance}
\usepackage{subcaption}
\usepackage{multirow}
\usepackage{multicol}
\usepackage{bbm}
\usepackage{color}
\usepackage{xcolor}
\usepackage{xspace}
\usepackage{amsmath}
\usepackage{amsfonts}
\usepackage{amsthm}
\usepackage{amssymb}
\usepackage{float}
\usepackage{tabularx}
\usepackage{siunitx}
\usepackage{enumitem}
\usepackage[T1]{fontenc}
\usepackage{xparse}
\usepackage{mdframed}
\usepackage{xcolor}
\usepackage{pifont}

\newcolumntype{Y}{>{\centering\arraybackslash}X}

\title{FederatedScope-LLM: \protect{\\}A Comprehensive Package for Fine-tuning Large Language Models in Federated Learning}

\author{
\!Weirui Kuang\thanks{Co-first authors.}~~, Bingchen Qian$^*$, Zitao Li, Daoyuan Chen, Dawei Gao, Xuchen Pan, Yuexiang Xie,\\
\AND
~Yaliang Li\thanks{Corresponding author, email: yaliang.li@alibaba-inc.com}~~, Bolin Ding, Jingren Zhou\\
\\
\quad Alibaba Group
}

\newcommand{\rmnum}[1]{(\romannumeral #1)}

\newcommand{\ours}{\textsc{FS-LLM}\xspace}
\newcommand{\model}{\textsc{LLM-AlgZoo}\xspace}
\newcommand{\trainer}{\textsc{LLM-Trainer}\xspace}
\newcommand{\databar}{-}
\newcommand{\benchmark}{\textsc{LLM-Benchmarks}\xspace}
\newcommand{\package}{package\xspace}
\newcommand{\paper}{paper\xspace}

\iclrfinalcopy
\begin{document}

\maketitle
\begin{abstract}
Large language models (LLMs) have demonstrated great capabilities in various natural language understanding and generation tasks. 
Platforms such as Hugging Face facilitate access and utilization of the pre-trained LLMs for different entities, ranging from computer science researchers to users with little machine learning background. 
Different entities can further improve the performance of those LLMs on their specific downstream tasks by fine-tuning LLMs.
When several entities have similar interested tasks, but their local data cannot be shared directly because of privacy concerns regulations, federated learning (FL) is a mainstream solution to leverage the data of different entities.
Besides avoiding direct data sharing, FL can also achieve rigorous data privacy protection, model intelligent property protection, and model customization via composition with different techniques.
However, fine-tuning LLMs in federated learning settings still lacks adequate support from the existing FL frameworks because it has to deal with optimizing the consumption of significant communication and computational resources, various data preparation for different tasks, and distinct information protection demands. 
This paper first discusses these challenges of federated fine-tuning LLMs in detail, and introduces
our implemented \package \textbf{F}ederated\textbf{S}cope-\textbf{LLM} (\ours) as a main contribution, which consists of the following components:
(1) we build a complete end-to-end benchmarking pipeline, automizing the processes of dataset preprocessing, federated fine-tuning execution or simulation, and performance evaluation on federated LLM fine-tuning with different capability demonstration purposes;
(2) we provide comprehensive and off-the-shelf federated parameter-efficient fine-tuning (PEFT) algorithm implementations and versatile programming interfaces for future extension to enhance the capabilities of LLMs in FL scenarios with low communication and computation costs, even without accessing the full model (e.g., closed-source LLMs);
(3) we adopt several accelerating operators and resource-efficient operators for fine-tuning LLMs with limited resources and the flexible pluggable sub-routines for interdisciplinary study (e.g., LLMs in personalized FL).
We conduct extensive and reproducible experiments to validate the effectiveness of \ours and benchmark advanced LLMs with state-of-the-art parameter-efficient fine-tuning algorithms in a federated setting, which also yields many valuable insights into federated fine-tuning LLMs for the research community.
To facilitate further research and adoption, we release \ours at \href{https://github.com/alibaba/FederatedScope/tree/llm}{https://github.com/alibaba/FederatedScope/tree/llm}. \footnote{We will continuously update the codebase and arXiv version.}
\end{abstract}

\newpage
\section{Introduction}
\label{sec:intro}
Recent advances in large language models (LLMs)~\citep{llama,gpt3,gpt4,opt,bloom,GLM,palm} have enabled a wide range of real-world applications across various domains, such as chatbots (i.e., ChatGPT\footnote{https://openai.com/blog/chatgpt}), writing assistants~\citep{CoAuthor,CreativeWriting}, search engines (i.e., New Bing\footnote{https://www.bing.com/new}), tool/API retriever~\citep{ToolLLM,Gorilla} and multimodal systems~\citep{mm1,mm2,mm3}. 
Compared to previous pre-trained language models~\citep{bert,RoBERTa}, LLMs exhibit remarkable emergent abilities that have not been observed before~\citep{zhao2023survey}.
These emergent abilities of LLMs are the cores of the unprecedented proficiency and efficiency of AI systems built on top of them.
Consequently, both academic and industrial people have demonstrated a keen interest in investigating the potentialities of LLMs.

When applying LLMs in practical applications, such as education, law, and medicine, fine-tuning LLMs with domain-specific data can be essential.
Fine-tuning can enrich LLMs with domain knowledge, enhance their specific ability, improve the fairness and reliability of the outputs, and prevent certain damage caused by hallucination~\citep{hallucination}.
However, fine-tuning LLMs entails a high demand for computational resources and a substantial amount of domain data that may not be sharable due to privacy concerns.
The former challenge can be addressed by recent works~\citep{lora,PrefixTuning,PTuningV2,ptuning,prompttuning,adapter,compacter,fusion,invertible}, which adapt pre-trained LLMs to specific domains by tuning modules with limited trainable parameters (denoted as \emph{adapters}). 
For the latter issue, one of the feasible solutions is federated learning (FL)~\citep{fl1,fl2,fl3}, a distributed learning paradigm that allows multiple entities to optimize a model collaboratively without directly sharing their data.

Although the existing FL frameworks~\citep{tff,pysyft} can usually support various machine learning models, 
the development of federated fine-tuning on LLM is still in a premature stage because of the following \textbf{gaps} in existing work.
\textbf{\rmnum{1}} No existing FL package contains comprehensive and efficient implementations of LLM fine-tuning algorithms and a standardized benchmark for comparing the model performance, communication cost, and computation overhead when federated fine-tuning LLMs.
\textbf{\rmnum{2}} Fine-tuning LLMs in FL is still computationally expensive on the client side, even with the parameter-efficient fine-tuning (PEFT) algorithms.
\textbf{\rmnum{3}} 
Because pre-trained LLMs are of great intelligent property value and may not belong to clients, it might be necessary to let clients conduct federated fine-tuning without accessing the full model (e.g., closed-source LLMs).
\textbf{\rmnum{4}} 
It is unclear whether the existing algorithms for solving advanced FL problems, such as personalized FL (pFL)~\citep{pFL,pFL-bench} and federated hyperparameter optimization (FedHPO)~\citep{FedHPOBench}, are still effective with different federated fine-tuning algorithms for LLMs.

We aim to bridge the aforementioned gaps and further promote the study of fine-tuning LLMs in the context of federated learning. Thus, we build up a novel open-source \package for fine-tuning LLMs via federated learning, called \textbf{F}ederated\textbf{S}cope-\textbf{LLM} (\ours), on top of FederatedScope (FS)~\citep{federatedscope}. 
Our contributions can be summarized as follows:
\begin{itemize}[leftmargin=0.50cm, itemindent=0cm]
    \item \ours packages a collection of diverse federated fine-tuning datasets from various domains with tunable levels of heterogeneity and a suite of corresponding evaluation tasks to form a complete pipeline to benchmark federated fine-tuning LLMs algorithms in FL scenarios.
    \item 
    \ours provides comprehensive federated fine-tuning algorithms for LLMs with low communication and computation costs and versatile programming interfaces, which support both scenarios where clients can or cannot access the full model.
    \item 
    \ours is equipped with an optimized federated fine-tuning training paradigm for LLMs towards customizable efficiency-boosting (e.g., memory consumption reduction and multi-GPU parallelism) and interdisciplinary research potentials (e.g., pFL and FedHPO).
    \item We perform extensive experiments based on \ours and investigate the empirical performances of federated fine-tuned LLMs. 
    Based on our observations, we point out the challenges for federated fine-tuning LLMs and offer insights for future research in this emerging field.
\end{itemize}
\begin{figure}[ht]
    \centering
    \includegraphics[width=0.95\textwidth]{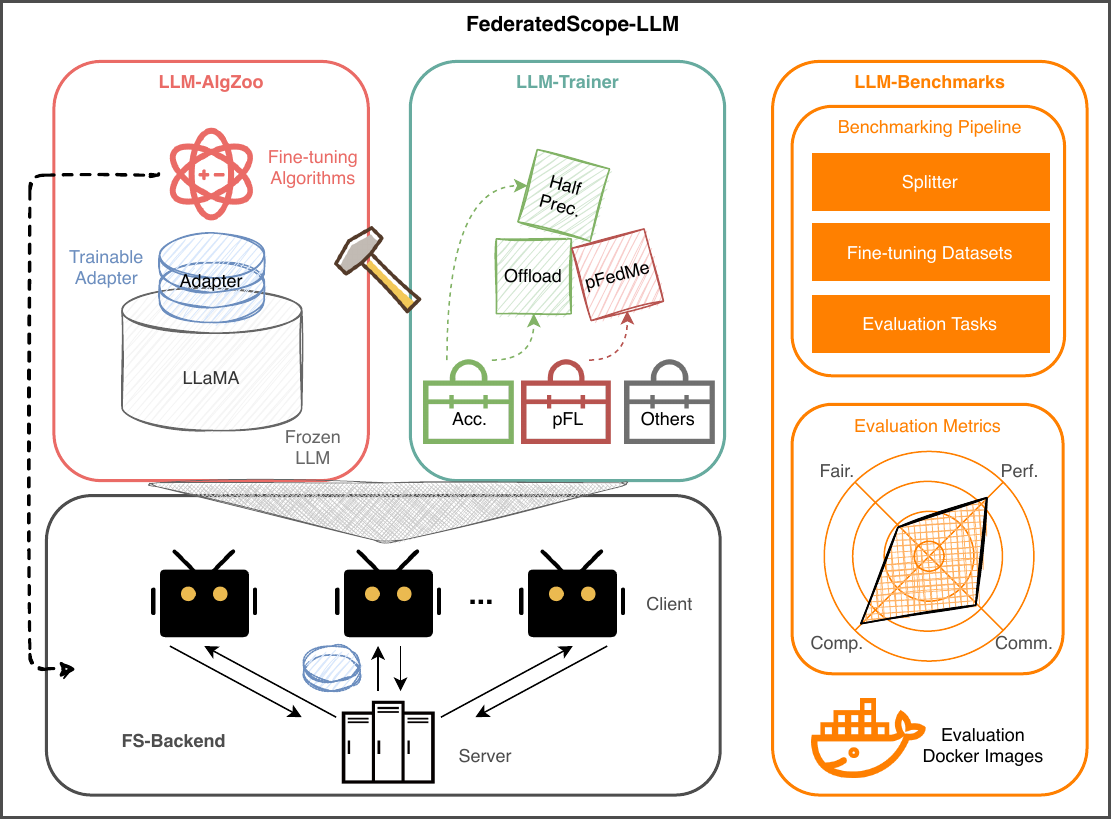}
    \caption{Overview of the architecture of \ours, which consists of three main modules: \benchmark, \model, and \trainer. As an example in the figure, we use the PEFT algorithms to fine-tune LLaMA~\citep{llama} in FL, with half-precision~\citep{mixprec} training and offloading~\citep{offloading} strategy and pFedMe~\citep{pFedME} algorithm. Under this learning paradigm, the clients can efficiently train on their local data with limited hardware resources, while the communication between the clients and the server only requires transmitting the adapter (which typically has very few parameters). This achieves high efficiency in both communication and computation. In the figure, Acc. stands for accelerating operator, Perf. stands for performance, Comm. stands for communication, Comp. stands for computation, and Fair. stands for fairness.}
    \label{fig:overview}
\end{figure}

\section{Overview}
\label{sec:framework}
In Figure~\ref{fig:overview}, we illustrate the overall architecture of \ours using a concrete example. 
As mentioned above, \ours is built upon FS~\citep{federatedscope}, an easy-to-use platform that provides essential modules for constructing FL courses (e.g., communication modules, aggregation strategies, and training functionality interfaces), rich extended functionality, and plenty of instantiated FL algorithms.
On top of FS, we develop three enrichments to address the gaps mentioned in Section~\ref{sec:intro} and facilitate fine-tuning LLMs in the federated setting: \textbf{\benchmark}, \textbf{\model}, and \textbf{\trainer}.
With these new modules, \ours supports end-to-end federated fine-tuning for LLMs, providing (1) data preparation for fine-tuning and evaluation, (2) several out-of-the-box popular fine-tuning algorithms and unified and flexible programming interfaces, (3) various accelerating and resource-efficient operators and flexible pluggable sub-routines for interdisciplinary study.
We present an overview in this section and give detailed descriptions of these modules in Section~\ref{sec:bench}, \ref{sec:model}, and \ref{sec:train}, respectively.

To consolidate the implementations and facilitate the FL research about LLMs, one of the critical steps is to construct a module with an end-to-end benchmarking pipeline, i.e., \benchmark in \ours.
We assemble various corpus datasets from different domains for fine-tuning and pair each of them with one specific relevant evaluation task to assess the performance of the fine-tuned LLMs on different domains. 
For the datasets prepared for fine-tuning, we partition them according to their meta-information, formulating them into federated versions. 
Meanwhile, \ours offers \textit{Splitter} to split centralized datasets, enhancing the extensibility of the fine-tuning dataset for federated fine-tuning LLMs. 
Furthermore, we provide a rich set of docker images where the runtime environment of the evaluation tasks has been prepared.
\benchmark enables users to conveniently compare the effectiveness of different fine-tuning algorithms in various FL scenarios.

Another important and helpful component provided in \ours, \model, includes a collection of fine-tuning algorithms tailored for FL.
As communication and computation resources of clients are usually limited in FL, we leverage and integrate several PEFT algorithms into the FL setting, such as LoRA~\citep{lora}, prefix-tuning~\citep{PrefixTuning}, P-tuning~\citep{ptuning}, and prompt tuning~\citep{prompttuning}.
Compared to full-parameter fine-tuning, these PEFT algorithms significantly reduce memory consumption, training time, and communication cost for fine-tuning LLMs.
Besides, motivated by the practical concerns on intelligent property protection of LLMs, we also integrate a privacy-preserving fine-tuning algorithm, offsite-tuning~\citep{offsite}, for the scenario where clients only tune small adapters based on a distilled model from a full LLM.

Based on the fine-tuning algorithms mentioned above (i.e., the PEFT algorithms and offsite-tuning), \trainer powers the FL fine-tuning process by adopting a hook-like training scheme.
Though these algorithms fine-tune a small number of parameters of LLMs, they can still be computationally expensive for some clients. 
Therefore, \ours incorporates various accelerating operators and resource-efficient operators as hook-like functions, such as DeepSpeed’s ZeRO~\citep{deepspeed}, Pytorch’s data parallelism~\citep{pytorch}, and model quantization~\citep{quant}, to accelerate the local fine-tuning process and enable \ours to work efficiently on consumer-level GPUs.
Besides, \trainer, with the help of its customizable hook-like programming interfaces, can be quickly and seamlessly integrated with existing plug-ins in FS for interdisciplinary studies, such as pFL~\citep{pFL} and FedHPO~\citep{FedHPOBench}, which could be a promising direction for further research.
For meeting different hardware settings or research goals, \trainer enables \ours to fine-tune LLMs in simulated mode (single-machine FL for simulation, all clients in one machine), distributed mode (multi-machine FL, one client per machine), and clustered mode (multi-machine FL, one client per cluster).

\section{\benchmark: A Complete Pipeline for Benchmarking}
\label{sec:bench}
As gap \textbf{\rmnum{1}} mentioned in Section~\ref{sec:intro}, there has yet to be a consensus in the academic community or industry on how to fairly evaluate the LLM fine-tuning algorithms in FL and what baselines are to compare. 
Therefore, we introduce \benchmark, the first convenient and fair module to evaluate federated LLM fine-tuning. 
\benchmark covers a complete benchmarking pipeline, consisting of stages from the construction of fine-tuning datasets to the evaluation with a collection of tasks.
To facilitate replication and validation, we offer a series of look-up tables containing benchmark results for the fine-tuning datasets and their corresponding evaluation tasks.
Additionally, we containerize the runtime environment of the evaluation for conveniently benchmarking the performance of the federated fine-tuned LLMs.

\subsection{Federated fine-tuning dataset construction}
\label{subsec:dataset}
Unlike pre-training from scratch, fine-tuning LLMs is typically for adapting pre-trained LLMs to one specific domain, which can be very diverse, such as code generation and mathematical reasoning. 
Additionally, as considered in existing FL literature, the local datasets held by individual clients can exhibit varying degrees of heterogeneity, even within the same domain. 
For instance, although a set of clients share a common interest in fine-tuning an LLM for code generation, the code base of some clients may mainly consist of Jave, while the code bases of other clients may contain a substantial portion of Python.

To echo the diversity of target domains and the heterogeneity of data in real-world FL scenarios, we curate three fine-tuning datasets in \benchmark. They cover a wide range of domains and exhibit realistic data distributions across different clients.
(1) \textit{Fed\databar CodeAlpaca} is built from CodeAlpaca~\citep{codealpaca} to enhance LLMs’ code generation capability.
It simulates a \emph{nine}-client FL scenario. 
Each client's dataset consists of coding exercises with answers limited to \emph{one} specific programming language, such as Java or Python. 
(2) \textit{Fed\databar Dolly} is for enhancing the capability of LLMs for generic language. We partition Databricks-dolly-15k~\citep{DatabricksBlog2023DollyV2} into \emph{eight} clients' local datasets.
Each client's dataset consists of a series of high-quality human-generated question-response pairs but is limited to \emph{one} specific NLP task, such as information extraction or QA.
(3) \textit{Fed\databar GSM8K-3} is curated into \emph{three} subsets from the train split in \textit{GSM8K}~\citep{cobbe2021gsm8k}, aiming to enhance the capability of LLMs for the chain of thought (CoT). 
Each client's dataset consists of grade school math questions randomly partitioned from the original dataset.

It is worth noting that, in addition to the built-in datasets mentioned above, we provide splitters for partitioning the centralized dataset into a federated version based on different meta-information or with different heterogeneity degrees among clients.
For example, users can apply different splitters to a centralized dataset, such as the uniform splitter, Dirichlet splitter, or meta-information splitter, to construct fine-tuning datasets that mirror the heterogeneity inherent in different FL scenarios. 
We provide several built-in datasets for these splitters, such as \textit{Alpaca}~\citep{alpaca}, \textit{cleanedAlpaca}~\citep{cleanedalpaca}, etc. 
For more details about the provided fine-tuning datasets, please refer to Appendix~\ref{app:data}.

\begin{table*}[ht]
\renewcommand{\arraystretch}{1.5}
\centering
\caption{Federated fine-tuning dataset and its corresponding evaluation task.}
\begin{tabular}{p{2.7cm}p{3.3cm}p{6.7cm}}
\toprule
 & Fine-tuning Dataset &  Goal of Evaluation Task \\ \hline
\makecell[tc]{\textit{Fed\databar CodeAlpaca} \\ \& \\ \textit{HumanEval}} & Coding exercises with different programming languages & How much can a federated fine-tuning algorithm improve the performance of an LLM with heterogeneous data on an in-domain task (coding)? \\ \hline
\makecell[tc]{\textit{Fed\databar Dolly} \\ \& \\ \textit{HELM}} & Human-generated question-response pairs with different types & How much can a federated fine-tuning algorithm improve the performance of an LLM with heterogeneous data on generic language capabilities?\\ \hline
\makecell[tc]{\textit{Fed\databar GSM8K-3} \\ \&    \\ \textit{GSM8K-test} } &  Math questions with independent and homogeneous distribution & How much can a federated fine-tuning algorithm improve the performance of an LLM with i.i.d. data on an in-domain task (CoT)? \\
\bottomrule
\end{tabular}
\label{tab:dataset-comb}
\end{table*}

\subsection{Federated LLM fine-tuning evaluation}
\label{subsec:eval}
LLMs are known to be very powerful, but it is challenging to evaluate their capabilities by a single metric. 
To the best of our knowledge, there are no ready-to-use evaluation tools for assessing federated LLM fine-tuning (let alone with the personalized FL algorithms) in terms of accuracy and efficiency.
To fulfill such a gap, in \benchmark, we provide a complete benchmarking pipeline to assess LLM fine-tuning in various FL scenarios.

We argue that fine-tuning should aim to enhance one of the two aspects of LLMs: either to improve their generic language capabilities or to improve their domain-specific capabilities for one particular downstream task.
Therefore, we curate three evaluation datasets from different subject areas, including 
\textit{HumanEval}~\citep{HumanEval} for the code generation, \textit{HELM}~\citep{helm} for the generic language capabilities, 
and \textit{GSM8K-test}~\citep{cobbe2021gsm8k} (the test split in \textit{GSM8K}) for CoT.
Given that different datasets employ different default evaluation metrics, for simplicity, we introduce the term \emph{evaluation score} as a unifying descriptor for the evaluated results obtained on these datasets with their metrics.
Specifically, the evaluation score represents Pass@1 score when using \textit{HumanEval}, a mixture of metric scores on 16 subtasks when evaluating on \textit{HELM}\footnote{Because evaluation on 16 subtasks is time-consuming, we also curate a smaller \textit{HELM-MINI} with 4 subtasks but consistent performance with original \textit{HELM} for a brief evaluation. }, and accuracy when utilizing \textit{GSM8K-test}.
And more details can be found in Appendix~\ref{app:eval}.

Then, we define evaluating an LLM on a specific dataset and generating the corresponding evaluation score as an \emph{evaluation task}.
We combine each fine-tuning dataset mentioned in Section~\ref{subsec:dataset} with one specific evaluation task, which allows users to fairly assess the improvement of fine-tuned LLMs in FL.
The fine-tuning dataset, the corresponding evaluation task, and the goal of the evaluation task are listed in Table~\ref{tab:dataset-comb}.
Note that there is generally a distribution shift between the fine-tuning and evaluation datasets, making it much more challenging than those in other domains of FL~\citep{leaf,dong2022collaborating,FS-GNN}.
To ensure the consistency of the evaluation results, we containerize the runtime environment of these evaluation tasks to docker images for conveniently assessing the performance of the federated LLM fine-tuning.

Last but not least, we also introduce a rich set of cost-related metrics to measure the efficiency of a federated fine-tuning process, including both computation costs (e.g., GPU usage, computation time, flops count) and communication costs (e.g., message size).
Along with evaluation scores, these metrics could give a comprehensive assessment of the federated fine-tuning process.

\section{\model: A Collection of Fine-tuning Algorithms}
\label{sec:model}
In addition to the benchmarking module, \benchmark, we implement a set of popular fine-tuning algorithms in \ours and introduce them as \model in this section.
Aiming to fulfill the gaps \textbf{\rmnum{1}} and \textbf{\rmnum{2}}, \model first includes a collection of PEFT algorithms to satisfy the constraints on the communication and computation costs in federated fine-tuning when all clients have access to the full model.
However, we also realize that there are cases where the LLM owner may not be willing to share the full model in the federated fine-tuning stage.
Thus, to fulfill gap \textbf{\rmnum{3}}, we further adopt a fine-tuning algorithm that does not require full model access in the FL setting.
Notice that all these fine-tuning algorithms are implemented on a set of unified but flexible interfaces, which hides the underlying standard functionalities from algorithm implementors, such as coordinating communication in FL.
The same interfaces can support more diverse fine-tuning algorithms in future development if users follow the same programming paradigm as our implemented algorithms.

\subsection{Reducing communication and computation costs with PEFT algorithms}
\label{subsec:adapter}
Achieving communication and computation efficiencies are two major challenges for fine-tuning LLMs in FL.
The communication bottleneck arises from the limited bandwidth of the internet connection between the server and the clients. 
This challenge becomes exacerbated when full-parameter fine-tuning LLMs in FL, as they require transmitting more parameters than previous pre-trained language models.
For example, full-parameter fine-tuning LLaMA-7B in FL requires 28GB of message transfer for one communication round from the client to the server. Assuming the network bandwidth is 100MBps, only model uploading and downloading will take about 75 minutes, which is intolerant, especially for clients with limited network bandwidth.
The computation efficiency is the other critical issue for federated fine-tuning LLMs.
For example, full-parameter fine-tuning LLaMA-7B requires about 28GB of GPU memory for the model. 
In addition, the SGD optimizer and the gradients need another 92GB of GPU memory, leading to at least 112GB of GPU memory in total - this is unaffordable for most resource-limited entities. 

We provide a set of implemented PEFT algorithms in \ours as solutions for our users to encounter these challenges, including LoRA~\citep{lora}, prefix-tuning~\citep{PrefixTuning}, P-tuning~\citep{ptuning}, and prompt tuning~\citep{prompttuning}.
These algorithms perform fine-tuning by only training (additional) modules with limited parameters, known as \emph{adapters}, but keep other parameters frozen.
Compared to full-parameter fine-tuning in FL, clients only need to transmit the adapters in each communication round, which reduces the transmission time to tens of seconds or even seconds.
Meanwhile, PEFT algorithms reduce the computation cost and make local training more viable for resource-limited clients. 
For example, if we only fine-tune the adapter of LLaMA-7B, the total GPU memory consumption will be a little more than 28GB, and the computation time will be less than that of full-parameter fine-tuning as well.
Thus, based on considerations of communication and computation costs, \model adopts PEFT algorithms and makes them considerably more viable for resource-limited entities when federated fine-tuning LLMs.

\subsection{Federated fine-tuning without accessing full model}
\label{subsec:offsite-tuning}
Many existing LLMs~\citep{gpt35,gpt4,claude} are closed-source for some reasons such as high training costs, preventing training data leakage, and maintaining commercial secrets. 
However, in some commercial scenarios, downstream customers may not be satisfied with simply using APIs to perform inference on these all-around LLMs but also want to customize them more to their domain-specific data.
These domain-specific data are often private, limited, and incomplete, which leads to insufficient adaptation and generalization of LLMs to the customers’ needs.

To satisfy such practical demand, we adapt a privacy-preserving fine-tuning algorithm, offsite-tuning~\citep{offsite}, to a federated version, and name it FedOT for short.
It sends a lossy compressed model with untrainable parameters to the clients as an emulator of the complete LLM at the beginning of FL.
During the FL, the clients fine-tune adapters with the frozen emulator and their domain-specific data.
FedOT safeguards both the intelligent property of the model providers and the data privacy of the clients, while leveraging the distributed data for adapting LLMs to specific domains.
This algorithm can be further integrated with the PEFT algorithms mentioned in Section~\ref{subsec:adapter}.

\subsection{Extendability by unified interfaces for federated fine-tuning}
\begin{figure}[ht]
    \centering
    \includegraphics[width=1.0\textwidth]{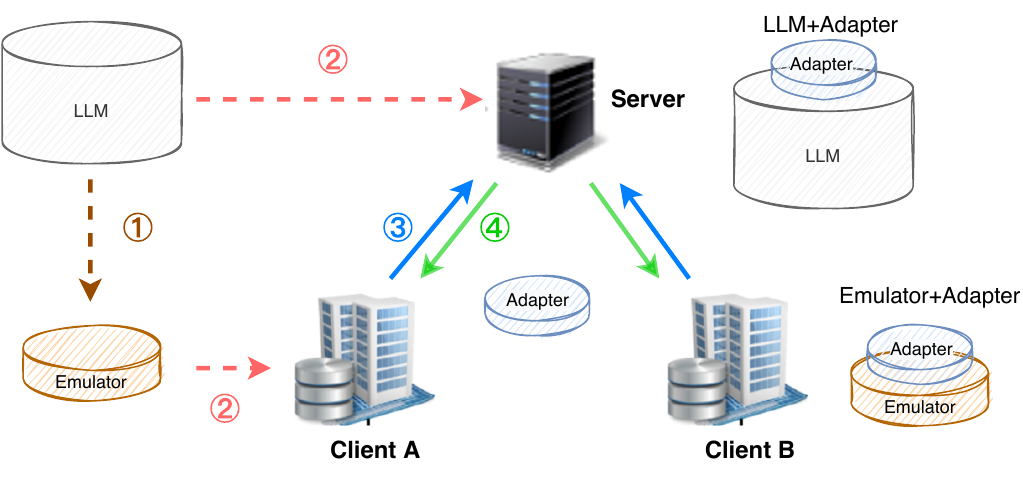}
    \caption{The unified interfaces for federated fine-tuning LLMs with or without accessing the full model. 
    When the LLM is not accessible to clients, different algorithms can be used to generate an emulator, including distillation, pruning, and quantization via \ding{172} LLM model pre-processing interface; if the LLM is accessible, \ding{172} just output the input by default. 
    The other three interfaces in the figure are \ding{173} initial model broadcast, \ding{174} shared parameter aggregation, and \ding{175} parameter re-distribution.}
    \label{fig:offsite-tuning}
\end{figure}

Notice that the aforementioned federated fine-tuning algorithms, regardless of whether the clients have access to the full model, are all implemented via a set of unified interfaces in \ours. 
The interfaces compose a skeleton structure that can be customized to various FL applications.
The interfaces can be invoked with customized functions in different stages, handling the underlying communication and synchronization between the server and clients in FL.
Figure~\ref{fig:offsite-tuning} illustrates some unified interfaces used by fine-tuning algorithms in Section~\ref{subsec:adapter} and \ref{subsec:offsite-tuning}.
The unified interfaces include but are not limited to the model pre-processing interface (arrow \ding{172}), initial model broadcast interface (arrow \ding{173}), shared parameter aggregation interface (arrow \ding{174}), and parameter re-distribution interface (arrow \ding{175}).
For closed-source LLMs, which are inaccessible to clients, the model providers can compress the LLM into an emulator by implementing a distillation function with the model pre-processing interface; for open-source LLMs, which are accessible, the pre-processing interface just returns the original LLM. 

These unified but flexible interfaces are based on the event-driven architecture in FL~\citep{federatedscope}. 
In the context of federated LLM fine-tuning, events are the exchanged information (e.g., local updated weights).
Each event has a corresponding handler, which is the actions triggered by it. 
The event-driven architecture allows the entities of FL to program their behaviors and react to the events.
Thus, by designing message-handler pairs, users can extend and customize the federated fine-tuning algorithms easily because of the extensibility of \ours.

\newpage
\section{\trainer: Training Operators and Paradigm}
\label{sec:train}
Although PEFT algorithms can significantly reduce the computation cost, they may still be computationally expensive for some clients with limited resources.
To alleviate such concerns, we provide \trainer, which is designed to further accelerate computation and save resources during the local training and message transmission stages in FL.
We implement a set of accelerating operators and resource-efficient operators to fulfill gap \textbf{\rmnum{2}}.
Moreover, to meet different hardware settings or research goals,
\ours supports simulated mode, distributed mode, and clustered mode. 
For simulated mode, all clients run on a single machine to simulate the federated fine-tuning process.
For the distributed mode and clustered mode, each client runs on one or more machines and communicates with the server separately.
These modes share a consistent programming paradigm and behavior.
Meanwhile, we note that there are also advanced FL problems in federated fine-tuning LLMs, such as pFL~\citep{pFL} and FedHPO~\citep{FedHPOBench}.
Thus, to further fulfill gap \textbf{\rmnum{4}}, we implement \trainer as a collection of hook-like functions to support rich extensions, following the design principles of FS Trainer~\citep{federatedscope}.
These hook-like functions will be executed by some arranged pattern to fine-tune the adapter within the local client.
By adding, removing, or replacing hook-like functions, entities can conveniently customize local fine-tuning procedures and extend applicabilities.
\trainer can be effortlessly compatible with implemented advanced FL algorithms for interdisciplinary research.

\subsection{Training operators for acceleration and efficiency}
\label{subsec:accelerator}
We introduce various accelerating operators and resource-efficient operators for fine-tuning LLMs in \ours. 
These provided operators aim to optimize the federated fine-tuning process in terms of CPU/GPU memory consumption, multi-GPU parallel, and communication cost. 
We describe the operators and show how they can be combined to achieve better compatibility and efficiency.

\noindent\textbf{Mode-generic operators.}
We provide operators generalized to different modes in the local fine-tuning stage of \ours. 
We implement mixed-precision training and gradient accumulation operators in several hook-like functions to save GPU resources. 
Furthermore, to accelerate the local fine-tuning process and enable multi-GPU parallelism, \ours integrates Pytorch’s data parallel mechanism. 

\noindent\textbf{Mode-specific operators.}
Besides the aforementioned generalized operators, we develop specialized operators that are tailored for each mode and aim to address the bottlenecks in each mode.
To be more specific, in the simulated mode, instantiating multiple clients on a single machine with several independent instantiated models causes a lot of memory consumption. 
Therefore, we use a round-robin switching operator, which allows the clients to take turns using the frozen full model for fine-tuning the adapter and then aggregating the updated adapters. 
Under this operator, when the number of clients grows, the memory consumption will only increase by an additional amount of the adapter.
This improvement makes it possible to conduct simulated FL experiments with a number of clients on one single machine.
For the distributed mode and clustered mode, we accelerate the communication between the server and the clients by one to two orders of magnitude by reducing the size of communication messages.
We apply different communication optimizations for different messages and introduce communication-efficient operators, including quantization operator, streaming operator, and compression operator.
Specifically, the quantization operator reduces the bit-width of the model parameters in the message to 16 or 8 bits; the streaming operator serializes the model parameters to eliminate the overhead of type conversion; the compression operator applies DEFLATE~\citep{deflate} or Gzip~\citep{gzip} algorithms to compress the messages.

\begin{figure}[ht]
    \centering
    \includegraphics[width=1.0\textwidth]{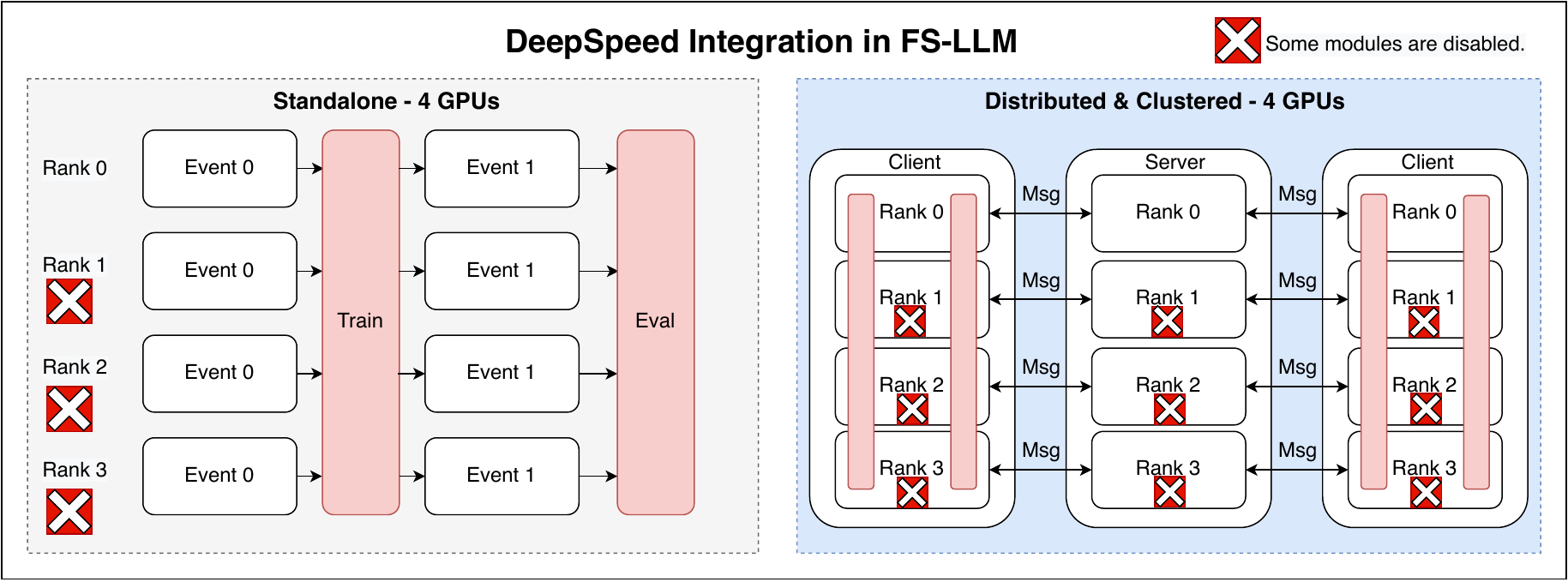}
    \caption{\ours integrates DeepSpeed for federated fine-tuning in different hardware conditions. Rank $0$ indicates the main process for multi-GPU training, and some modules of other subprocesses are disabled (e.g., logging and saving checkpoints). Msg stands for messages transmitted between the server and the clients, which trigger the events to happen.}
    \label{fig:deepspeed}
\end{figure}

\noindent\textbf{Parallelization operators.}
Meanwhile, we also migrate the functionality of DeepSpeed as shown in Figure~\ref{fig:deepspeed}, providing data parallelism with multi-GPU memory optimization and CPU offloading capabilities, further enhancing resource utilization.
Specifically, after launching the fine-tuning with DeepSpeed, we disable some modules (e.g., the logging module, the WandB module, and the file writing module) for the subprocess other than the Rank $0$ process to avoid conflicts. 
In addition, in distributed and clustered modes, each subprocess communicates independently with each other across the server and the clients, and only synchronizes with other subprocesses during the local fine-tuning process, ensuring consistent behavior across different modes.

\subsection{Training paradigm for extendability towards advanced FL tasks}
\label{subsec:plugin}
Another design philosophy of \trainer is to support various extensions easily and integrate different hook-like functions with existing training paradigms seamlessly for FL interdisciplinary study. 
A vanilla fine-tuning process involves three steps: 
(1) preparing datasets and extracting them by batches, (2) iterating over the training datasets to update the model parameters, and (3) evaluating the performance of the fine-tuned model on validation datasets.
We implement fine-grained learning behaviors for advanced FL tasks at different points in the above steps with hook-like functions.
For instance, pFL~\citep{pFL} and FedHPO~\citep{FedHPOBench} are two advanced FL tasks that can significantly improve model performance in FL.

\noindent\textbf{Adaptation of pFL with LLMs.}
FS~\citep{federatedscope} provides many implemented pFL plugins~\citep{pFL-bench}, which can be integrated for federated fine-finetuning LLMs. 
However, due to limited resources, it is unrealistic for the clients to maintain both global and local models at the same time for some pFL algorithms~\citep{pFedME,ditto}, as it would consume a lot of memory. 
Therefore, we optimize the implementation of these pFL algorithms by only using global and local adapters. 
With such an implementation, by adding a pFL hook-like function, we can achieve the cooperation of any PEFT algorithms with the implemented pFL algorithms, which provides strong extensibility for federated fine-tuning LLMs in personalized settings.

\noindent\textbf{Adaptation of FedHPO with LLMs.}
Similarly, for FedHPO, we offer extensions for \ours with model-free HPO methods (e.g., random search~\citep{randomsearch} and grid search), model-based HPO methods (e.g., Bayesian Optimization~\citep{bayesianoptimization}), multi-fidelity HPO method (e.g., Successive Halving Algorithm~\citep{successivehalvingalgorithm}), and FedHPO methods (e.g., FTS~\citep{FTS} and FLoRA~\citep{flora}). 

As our vision, the extensibility of \ours to support the interdisciplinary study of LLMs and other FL scenarios dramatically expands the application scenarios of fine-tuning LLMs in FL and raises many new research opportunities, which we will discuss more details later in Section~\ref{sec:discussion}.

\section{Experiments}
\label{sec:exp}
In this section, we demonstrate the effectiveness of \ours by a set of comprehensive experiments with different algorithms and tasks. 
We want to answer the following research questions: 
(1) How effective and efficient is it to federated fine-tuning LLMs with PEFT algorithms (Section~\ref{subsec:main_exp} and \ref{subsec:gpu})? 
(2) How effective is it to federated fine-tune LLMs without accessing the full model (Section~\ref{subsec:offsite_exp})?
(3) What insights can we obtain from the interdisciplinary capabilities of \ours in resolving pFL and FedHPO problems when federated fine-tuning LLMs (Section~\ref{subsec:pfl} and \ref{subsec:fedhpo})? 
We go through our experimental results and provide answers to the above questions.

\subsection{Effectiveness of PEFT algorithms in \ours}
\label{subsec:main_exp}
Firstly, we benchmark the performance of different PEFT algorithms in different application domains and scenarios.
As described in Section~\ref{sec:bench}, we use three federated fine-tuning datasets to fine-tune LLMs and evaluate them with corresponding tasks: \rmnum{1} federated fine-tuning with \textit{Fed\databar CodeAlpaca} for code generation and evaluating with \textit{HumanEval}, \rmnum{2} federated fine-tuning with \textit{Fed\databar Dolly} for generic language capability and evaluating with \textit{HELM}, and \rmnum{3} federated fine-tuning with \textit{Fed\databar GSM8K-3} for mathematical reasoning and evaluating with \textit{GSM8K-test}.
We conduct experiments in three scenarios: global (centralized fine-tuning), fed (federated fine-tuning), and local (separated fine-tuning). 
To be more specific, the global scenario can be regarded as fine-tuning LLMs with one client who holds the whole fine-tuning dataset. Fed scenario means that clients federated fine-tune LLMs where each client holds a different fine-tuning dataset. 
Local scenario means that each client independently fine-tunes LLMs with its own fine-tuning dataset. 

All the experiments are conducted on the machines with the same hardware configuration: Nvidia A100 GPU (80GB) with Intel Xeon Platinum 8369B CPU and 512GB of RAM.
For all scenarios, we repeat the experiments three times with different random seeds.
We report the average evaluation score with its standard deviation. 

\noindent\underline{\textbf{Benchmark federated fine-tuned LLaMA-7B.}}
We use a widely adopted LLM, LLaMA-7B, with three PEFT algorithms~\footnote{
We exclude prefix-tuning from our experiments, because its implementation in ~\cite{peft} contains unresolved issues when we were preparing this package.
}, including LoRA~\citep{lora}, P-tuning~\citep{ptuning}, and prompt tuning~\citep{prompttuning}. 
We employ FedAvg~\citep{fl2} as the federated aggregation strategy.
To conduct the experiments uniformly and fairly, we fix the FL-specific hyperparameters and the hyperparameters that have a large impact on the computation cost for all experiments. 
For example, we set the communication round to \num{500}, the local update step to \num{30}, and the batch size to \num{1}.
We perform a grid search for algorithm-specific and learning-related hyperparameters to obtain the optimal configuration. 
For example, the search space of the learning rate is $\{\num{1e-4}, \num{3e-4}, \num{5e-4}, \num{1e-3}, \num{3e-3}, \num{5e-3}\}$.
Please refer to Appendix~\ref{app:details} for more algorithm-specific hyperparameters corresponding to each PEFT algorithm.
Moreover, to further reduce the GPU memory consumption for efficiency, we employ the half-precision operator during fine-tuning.

\begin{table*}[htbp]
\centering
\caption{Performance comparisons among different PEFT algorithms when fine-tuning LLaMA-7B in FL: Evaluation Scores(\%) $\pm$ standard deviation(\%).}
\label{tab:main_exp}
\renewcommand{\arraystretch}{1.5}
\begin{tabularx}{\textwidth}{cYYYYYY}
\toprule
Algorithm & Scenario  & \textit{Fed\databar CodeAlpaca} & \textit{Fed\databar Dolly} & \textit{Fed\databar GSM8K-3} \\ \hline
 \multirow{3}{*}{LoRA}
 & Global  & 13.54$\pm$0.24 &  46.25$\pm$0.44 & 14.81$\pm$1.04\\
 & Fed  & 13.29$\pm$0.10 & 46.57$\pm$0.24 & 14.25$\pm$1.37  \\ 
 & Local  & 10.99$\pm$0.77 & 43.98$\pm$1.38 & 11.88$\pm$1.35 \\ \hline
 \multirow{3}{*}{P-tuning}
 & Global & 10.24$\pm$0.30 & 41.29$\pm$0.01 & 12.13$\pm$0.41  \\ 
 & Fed & 9.71$\pm$0.66 & 41.50$\pm$0.32 & 11.75$\pm$0.39  \\ 
 & Local  & 7.78$\pm$2.27 & 38.76$\pm$2.39 & 11.42$\pm$0.96  \\ \hline 
 \multirow{3}{*}{Prompt tuning}
 & Global & 9.80$\pm$1.79 & 41.24$\pm$0.54 & 9.75$\pm$1.49 \\ 
 & Fed  & 9.63$\pm$0.36 & 40.72$\pm$0.64 & 9.86$\pm$0.59 \\ 
 & Local & 7.18$\pm$2.17 & 37.65$\pm$6.12 & 9.65$\pm$0.77   \\
\toprule
\end{tabularx}
\end{table*}

\noindent\textbf{Results and Analysis.}
Table~\ref{tab:main_exp} shows the comparisons among different PEFT algorithms for federated fine-tuned LLaMA-7B under different scenarios.
In summary, we can draw the following conclusions. 
(1) All algorithms with federated fine-tuning can significantly outperform those under the local scenario, and they all show very competitive results against those under the global scenario. 
This suggests that it is feasible and effective to federated fine-tuning LLMs with PEFT algorithms, which allows multiple entities to benefit from the collaborative training without directly sharing their private data.
(2) Among these PEFT algorithms, LoRA shows the most promising performance and beats the other two algorithms by a large margin in all three scenarios. 
P-tuning and prompt tuning are two parameterized prompt algorithms that insert some learnable tokens into the model input to improve the performance of downstream tasks.
However, they are outperformed by LoRA, a low-rank adaptation algorithm that augments each layer of LLMs with two low-rank matrices to capture new domain-specific knowledge.
We conjecture that these parameterized prompt algorithms are limited by the knowledge encoded in the original LLM during the pre-training stage.
This indicates that LoRA is a suitable PEFT algorithm for federate fine-tuning LLMs in future research or realistic scenarios.

\noindent\underline{\textbf{How about federated fine-tuning previous-generation language models?}}
As a comparison, we select the superior PEFT algorithm in Table~\ref{tab:main_exp}, LoRA~\citep{lora}, to federated fine-tune a previous-generation language model, OPT-2.7B~\citep{opt}, which has fewer parameters than LLaMA-7B. 
Table~\ref{tab:opt_exp} shows the evaluation results of OPT-2.7B on the \textit{HumanEval}, \textit{HELM} evaluations, and \textit{GSM8K-test}.

\begin{figure}[htbp]
    \centering
    \includegraphics[width=0.8\textwidth]{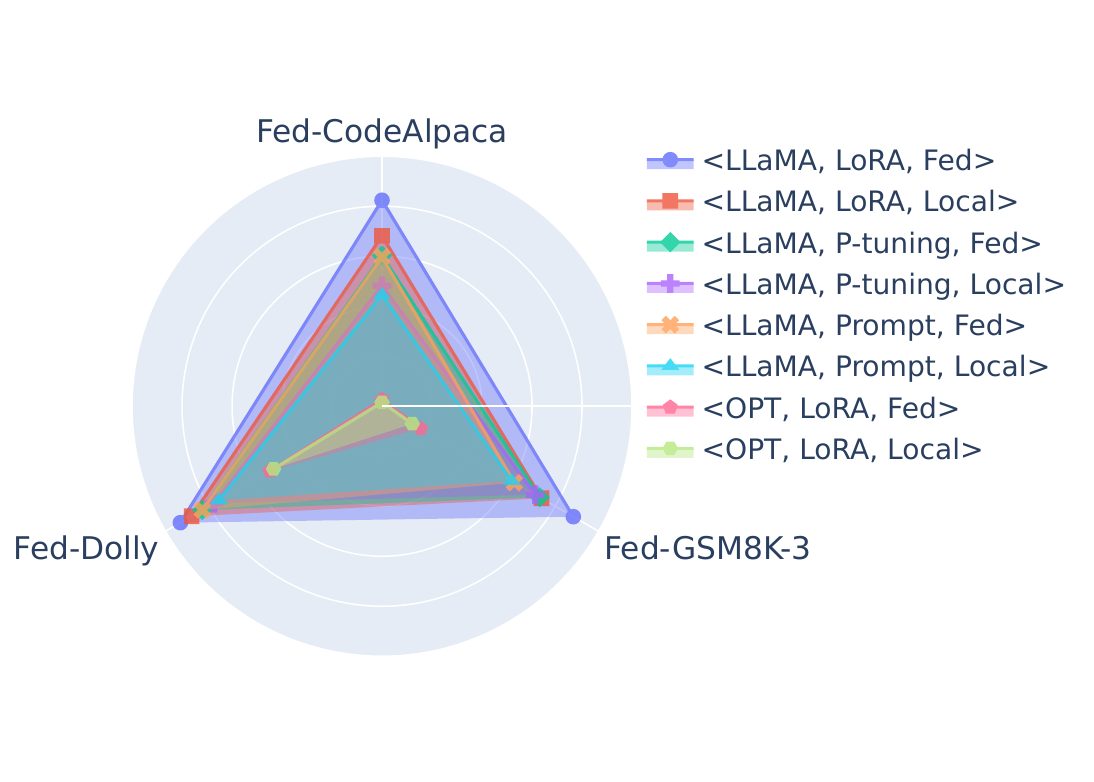}
    \vspace{-0.4in}
    \caption{Visualization of the performance comparison of fine-tuned LLaMA-7B and OPT-2.7B under federated and local scenarios. (The axes are scaled to highlight the differences.)}
    \label{fig:radar}
\end{figure}

Though the model performance of OPT-2.7B is improved with federated fine-tuning compared to those under the local scenario, it is not significant.
Moreover, OPT-2.7B fails in the evaluation of some subtasks in HELM due to the exceeded input length, which limits the scope of the application of the previous-generation language models with fewer parameters. 
Comparing the performance of federated fine-tuned LLaMA-7B and OPT-2.7B in Figure~\ref{fig:radar}, we observe that the current-generation LLM with a larger scale has an obvious advantage over the previous-generation language model with a smaller scale on various evaluation tasks. 
Thus, if the communication and computation costs are affordable, one should consider the larger scale of current-generation LLMs for more significant FL improvement and better absolute performance on downstream tasks.

\begin{table*}[ht]
\renewcommand{\arraystretch}{1.5}
\centering
\caption{Performance of federated fine-tuned previous-generation language model, OPT-2.7B, with LoRA: Evaluation Scores(\%) $\pm$ standard deviation(\%). ($^*$OPT-2.7B fails on some subtasks in HELM due to the exceeded input length, and failed subtasks are excluded when calculating the final results.)}
\label{tab:opt_exp}
\renewcommand{\arraystretch}{1.5}
\begin{tabularx}{\textwidth}{cYYYYYY}
\toprule
Algorithm & Scenario & \textit{Fed\databar CodeAlpaca} & \textit{Fed\databar Dolly} & \textit{Fed\databar GSM8K-3} \\ \hline
  \multirow{3}{*}{LoRA}
 & Global  & 0.61$\pm$0.61 & 25.90$\pm$0.40$^*$ & 2.92$\pm$0.11\\
 & Fed  & 0.43$\pm$0.07 & 25.90$\pm$0.33$^*$  & 2.88$\pm$0.17 \\
 & Local  & 0.25$\pm$0.25 & 25.06$\pm$1.02$^*$ & 2.25$\pm$0.22 \\
\bottomrule
\end{tabularx}
\end{table*}

\subsection{Efficiency of PEFT algorithms in \ours}
\label{subsec:gpu}
In this section, we evaluate the efficiency of various PEFT algorithms in FL. 
Among all the metrics, we focus on their GPU memory consumption, message size, and computation time in the federated fine-tuning process. 
We note that the GPU memory consumption refers to the model only, excluding the input tokens and the optimizer state, because the input length varies greatly across data and affects the GPU memory consumption during fine-tuning LLMs. 
For message size, we report the number of bytes of the serialized adapter parameters during one communication between the server and one client.
The computation time is defined as the time duration of one training step with a batch size of $1$, from the start of forward propagation to the end of backward propagation. 
We report the computation time on two types of hardware, both with 512GB of RAM: Nvidia A100 GPU (80GB) with Intel Xeon Platinum 8369B CPU and Nvidia V100 GPU (32GB) with Intel Xeon Platinum 8163 CPU.
For a fair comparison, we follow the same experimental settings as Section~\ref{subsec:main_exp} for all PEFT algorithms.

\noindent\textbf{Results and Analysis.}
As shown in Table~\ref{tab:efficiency}, we can first notice that (1) fine-tuning with different PEFT algorithms has negligible impact on the GPU memory consumption.
(2) However, there are large differences in the message sizes, which in turn lead to large variations in the transmission time.
For example, for a client with 100Mbps bandwidth, the transmission time (upload and download) per round ranges from about 0.01 seconds (with prompt tuning) to 40 seconds (with P-tuning) when using different PEFT algorithms. 
(3) Moreover, from the table, we can observe that the computation time varies due to different GPUs, with almost a twofold difference.
Therefore, a more critical issue deserves attention: federated fine-tuning LLMs may suffer from more idle time due to the heterogeneity of computing resources among different clients.
Because this computation efficiency difference can be significant, the benefit of mitigating communication latency by sending messages asynchronously in multiple computational stages may diminish. 

Therefore, there are two research directions for federated fine-tuning LLMs in efficiency: (1) how to leverage the idle time of computation-resource-rich clients while they wait for computation-resource-limited clients to complete local updates, and (2) how to optimize the utilization of the available bandwidth resources in computation-resource-limited clients during computation.

\begin{table*}[ht]
\centering
\caption{Efficiency comparisons among different PEFT algorithms when fine-tuning LLaMA-7B in FL. 
($^*$The GPU usage is subject to minor variations due to the differences in Cuda versions.)}
\label{tab:efficiency}
\renewcommand{\arraystretch}{1.5}
\begin{tabularx}{\textwidth}{cYYYYYY}
\toprule
  & LoRA & P-tuning & Prompt tuning \\ \hline
 GPU Usage$^*$ (MB)  & 13,450 & 13,538 & 13,442 \\ \hline
 Message Size (MB)  & 21.40 & 256.48 & 0.17 \\ \hline 
 Comp. Time on A100 (Sec.)  & 0.16$\pm$0.02 & 0.15$\pm$0.03 & 0.15$\pm$0.04 \\ \hline 
 Comp. Time on V100 (Sec.) & 0.33$\pm$0.07 & 0.33$\pm$0.08 & 0.33$\pm$0.10 \\
\bottomrule
\end{tabularx}
\end{table*}

\subsection{Fine-tuning LLMs without accessing the full model in FL}
\label{subsec:offsite_exp}
In this section, we investigate the performance of federated fine-tuning LLMs without accessing the full model. As mentioned in Section~\ref{subsec:offsite-tuning}, we adapt a privacy-preserving fine-tuning algorithm, offsite-tuning~\citep{offsite}, to federated scenarios. 
Specifically, we use the first and last two layers of LLaMA-7B as the adapter and compress the model as the emulator by dropping $20\%$ and $50\%$ of the remaining layers uniformly. 
Then the server broadcasts both the adapter and emulator to all clients, and the clients only fine-tune the adapter with FedAvg.
We compare the performance of LLMs with fine-tuned adapters via federated offsite-tuning (denoted as FedOT) and corresponding local offsite-tuning (denoted as LocalOT).
Following Section~\ref{subsec:main_exp}, we benchmark FedOT on \benchmark.

\noindent\textbf{Results and Analysis.}
We present the evaluation scores of FedOT and LocalOT in Table~\ref{tab:fedot}.
We can have the following observations:
(1) Comparing FedOT and LocalOT, FL offers significant benefits for this privacy-preserving fine-tuning algorithm for federated fine-tuning LLMs without accessing the full model.
This demonstrates that FedOT can still enable multiple entities to benefit from collaborative training without sharing their private data directly when they cannot access the full model.
(2) When the dropping rate is $20\%$, FedOT still achieves competitive performance compared to some of those PEFT algorithms, even though the clients cannot access the full model. 
However, it should be noted that this is because the number of parameters of the adapter in FedOT is significantly larger than those in the PEFT algorithms. 
FedOT sacrifices communication efficiency for model performance, making it effective in scenarios where clients cannot access the full model.
(3) On the other hand, the results show that when the dropping rate increases to $50\%$ from $20\%$, the model loses a large amount of knowledge from the pre-training stage, almost fails to retain the capacity of the CoT and code generation, and hardly acquire new knowledge from the fine-tuning.
There is a trade-off between the compression rate and the performance of LLMs: increasing the compression rate enhances the privacy of LLMs but degrades their performance.
This indicates that how to compress LLMs while maintaining their generalization ability and model privacy is a promising research direction to explore.

\begin{table*}[ht]
\centering
\caption{Performance comparisons between different compression rates (dropping layers uniformly) when fine-tuning LLaMA-7B without accessing the full model under federated and local scenarios: Evaluation Scores(\%) $\pm$ standard deviation(\%).}
\label{tab:fedot}
\renewcommand{\arraystretch}{1.5}
\begin{tabularx}{\textwidth}{cYYYYYY}
\toprule
Dropping Rate & Scenario & \textit{Fed\databar CodeAlpaca} & \textit{Fed\databar Dolly} & \textit{Fed\databar GSM8K-3}  \\ \hline
 \multirow{2}{*}{$20\%$}
 & Fed & 7.14$\pm$2.75 & 44.88$\pm$0.75 & 9.02$\pm$0.71   \\
 & Local  & 0.18$\pm$0.50 & 38.45$\pm$9.57 & 4.72$\pm$2.91 \\ \hline
 \multirow{2}{*}{$50\%$} 
 & Fed& 0.16$\pm$0.15 & 37.01$\pm$2.34 & 2.98$\pm$0.98 \\
 & Local & 0.00$\pm$0.00 & 35.44$\pm$5.99 & 1.82$\pm$1.29  \\
\bottomrule
\end{tabularx}
\end{table*}

\subsection{Personalized federated learning for LLMs}
\label{subsec:pfl}
To explore the potential of personalized federated learning algorithms~\citep{pFL} for fine-tuning LLMs in FL, we compare pFedMe~\citep{pFedME} and FedAvg~\citep{fl2} with LoRA in this subsection under different data heterogeneity. 
To simulate different levels of data heterogeneity, we create variants of \textit{Fed\databar CodeAlpaca}.

We use the programming language of the code samples as the label and split the fine-tuning dataset into nine clients by Latent Dirichlet Allocation (i.e., Dirichlet splitter).
We use four different values for $\alpha$ of the Dirichlet splitter from $0.05$ to $50.0$. 
Along with the original \textit{Fed\databar CodeAlpaca} dataset, we obtain five federated fine-tuning datasets with different heterogeneity.
Following Section~\ref{subsec:main_exp}, we repeat each experiment three times with different random seeds. 
The averaged evaluation scores (Pass@1 scores) with their standard deviation are reported. 
We note that evaluation scores with pFedMe are obtained by benchmarking each personalized client individually and then computing their average scores.

\begin{figure*}[ht]
	\centering
	\begin{subfigure}{0.48\linewidth}
		\centering
		\includegraphics[width=1.0\linewidth]{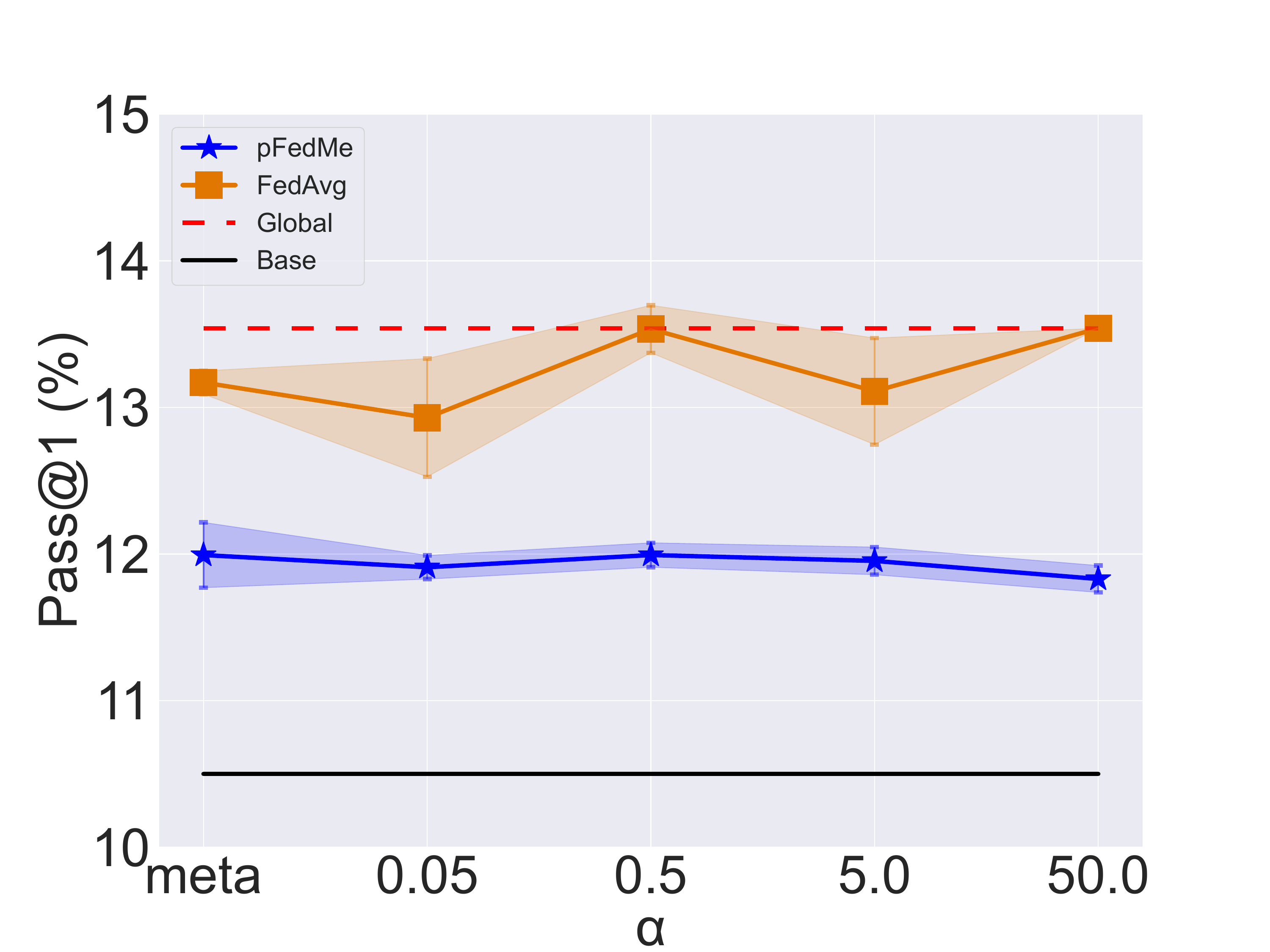}
		\caption{The comparison of performance with pFedMe and FedAvg over different data heterogeneity: the higher $\alpha$, the lower the data heterogeneity among clients. Global stands for fine-tuning LLMs under the global scenario, and Base stands for the original model, which is not fine-tuned.}
		\label{fig:pfl}
	\end{subfigure} \hspace{0.1in}
	\begin{subfigure}{0.49\linewidth}
		\centering
		\includegraphics[width=1.0\linewidth]{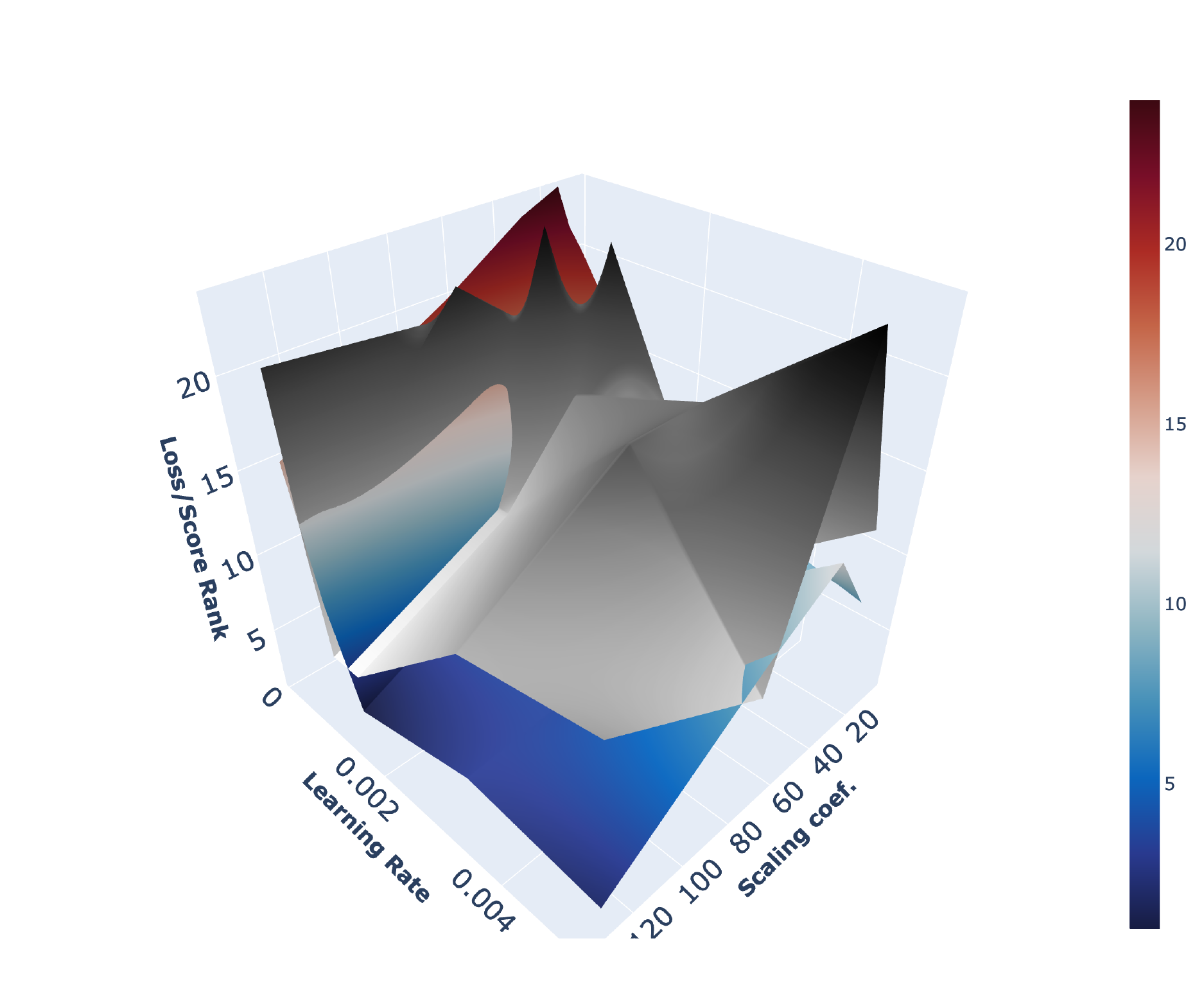}
		\caption{The landscape of the rank of the validation loss and the Pass@1 scores over all the hyperparameter combinations. The grey surface shows the distribution of the rank based on the Pass@1 scores, and the blue-red surface indicates the distribution of the rank based on the validation loss.}
		\label{fig:hpo}
	\end{subfigure}
	\caption{Fine-tuning LLMs in pFL (\textbf{Left}) and FedHPO (\textbf{Right}).}
	\label{fig:cross}
\end{figure*}

\noindent\textbf{Results and Analysis.}
We present the performance of fine-tuning with LoRA using FedAvg and pFedMe in different data heterogeneities in Figure~\ref{fig:pfl}, which shows the performance of fine-tuning with FedAvg gradually approaches that under the global scenario as data heterogeneity decreases.
However, pFedMe surprisingly does not outperform FedAvg under any data distribution, which shows different results with previous language models in pFL-bench~\citep{pFL-bench}.
We analyze this phenomenon and find that:
(1) To improve efficiency, we use the half-precision operator to fine-tune LLMs in the experiment. However, this leads to a more pronounced precision loss for pFedMe than FedAvg, since pFedMe multiplies the update of the local model with respect to the global model by a small learning rate. This adversely impacts the performance of pFedMe.
(2) The use of acceleration operators for efficient LLM fine-tuning restricts the range of hyperparameter space in these pFL algorithms, affecting the upper bound of algorithm performance in the valid parameter space. For example, assuming the LLM performs better when the learning rate is $0.00001$, but at this point, using half-precision or mixed-precision training for efficiency, the model can barely be updated due to the precision loss.

Based on these observations, we believe how to ensure the compatibility of various efficient training operators and different pFL algorithms is still unclear and deserves more attention from the community.
Besides, sharing a common base LLM and only maintaining multiple versions of adapters may not be compatible with some existing pFL algorithms because pFL algorithms may introduce access conflicts. 
For instance, when a penalty term is used to constrain the large updates of the local model, the computation to compute updates requires using both the global adapter with the base LLM and the local adapter with the base LLM in the same step.
Significantly, more memory cost is required to avoid the conflict by maintaining multiple copies of LLMs and their adapters.
Resolving this challenge may require new algorithm development or a new pFL computation pattern for future work.

\subsection{Study of the FedHPO for LLMs}
\label{subsec:fedhpo}
Since fine-tuning LLMs in FL is very costly, it is usually infeasible to perform full-fidelity hyperparameter optimization. 
However, we observe that the performance of fine-tuned LLMs in FL is highly dependent on the choice of hyperparameters (see Appendix~\ref{app:details}). 
Therefore, we investigate whether we can use low-fidelity FedHPO~\citep{FedHPOBench} methods in this scenario. 
We follow the experiment settings in Section~\ref{subsec:main_exp} and use LoRA to fine-tune LLaMA-7B on \textit{Fed\databar CodeAlpaca} yet with lower fidelity (i.e., fewer communication rounds). 
We rank all the hyperparameter combinations searched by their validation loss in ascending order and evaluation scores in descending order separately. 
We plot the two landscapes of the rank to explore the feasibility of using low-fidelity FedHPO methods when federated fine-tuning LLMs.

\noindent\textbf{Results and Analysis.}
Based on the results shown in Figure~\ref{fig:hpo}, we distill the following observations.
(1) We observe that the rank of the evaluation scores of the fine-tuned LLMs varies drastically and non-smoothly with respect to the hyperparameter changes. 
This poses a great challenge for finding the optimal hyperparameters, as it requires a fine-grained and exhaustive search over the hyperparameter space.
(2) Moreover, we reveal a significant discrepancy between the ranks of validation loss and the ranks of final generalization performance in evaluation tasks during fine-tuning. 
This implies that the validation loss may not reflect the generalization ability of the fine-tuned LLMs. 

In summary, we uncover two major challenges for fine-tuning LLMs in FedHPO: (1) the evaluation scores are highly sensitive and non-smooth to the hyperparameter changes, and (2) the validation loss may not be a reliable indicator of the generalization performance.
These challenges identify two promising yet unexplored research directions for future work on fine-tuning LLMs in FedHPO. 
The first direction is to develop fine-grained but efficient FedHPO methods for finding the optimal hyperparameters on federated fine-tuning LLMs, which can avoid exhaustive searches over the hyperparameter space. 
The second direction is to exploit concurrent exploration in FedHPO to evaluate the generalization ability of the client-side hyperparameters with low fidelity for each client during the FL process.

\section{Discussions and Future Directions}
\label{sec:discussion}
This paper introduces a comprehensive and practical \package for federated fine-tuning LLMs. 
Our experimental results illustrate how our \ours bridges the gaps between the universal FL framework and the need for fine-tuning LLMs under various FL settings.
More importantly, our benchmark results also provide positive guidance and insights for the research community regarding how to optimize the federated fine-tuning and what sub-problems in the field deserve more focus.

However, the results of this \paper are limited by several factors.
(1) Due to the resource limit, all experiments use a batch size of $1$, but federated fine-tuning LLMs with larger batch sizes might perform better.
(2) We also find that designing different prompts (either in fine-tuning or evaluation) will impact the evaluation results. To ensure a fair evaluation and comparison, we use a fixed prompt, but more explorations are possible.

Based on our observations and experiments, we outline some promising directions for future research in federated fine-tuning LLMs as follows.
\begin{itemize}[leftmargin=0.50cm, itemindent=0cm]
    \item Designing computation-efficient fine-tuning algorithms for federated fine-tuning LLMs. 
    Even with PEFT algorithms, the computation cost is still too high for most resource-limited clients. 
    Reducing the computation cost can lower the barrier for more data holders and allow more entities to benefit from the federated fine-tuning LLMs.
    \item Exploring more privacy-preserving fine-tuning algorithms without accessing the full model in FL. 
    FedOT suffers from a trade-off between model compression rate and model performance.
    Addressing this issue would protect the sensitive information of LLMs in FL from exposing pre-training data and the valuable full model, which could be exploited by malicious entities for adversarial attacks or commercial gains while maintaining model performance in FL.
    \item Optimizing pFL algorithms to enable robust combination with various accelerating and resource-efficient operators.
    If performance degradation due to low-precision training can be overcome, it would improve the personalized model performance when data are heterogeneous and computation resources are limited among clients.
    \item Investigating low-fidelity FedHPO methods for fine-tuning LLMs in FL. Based on our experimental results, we find the inconsistency between validation loss and the generalization performance of LLMs. 
    Overcoming this inconsistency would help find optimal hyperparameters for federated fine-tuning LLMs with low cost, resulting in better generalization performance.
    \item Extending the federated LLM fine-tuning to cross-device setting.
    As we have already observed the demand for federated fine-tuning LLMs in the cross-silo scenario, we also notice a similar need in the cross-device scenario~\citep{Lai2021FedScaleBM, chen2023fsreal, gao2023fsreal}. 
    In the cross-device scenario, the clients are more numerous and heterogeneous, the computational resources are more limited, and the network conditions are more diverse. 
    How to federated fine-tune LLMs under the cross-device scenario is an urgent problem that needs to be solved.
\end{itemize}
\section{Conclusions}
\label{sec:conclusion}
In this paper, we first identify gaps that need to be addressed between fine-tuning LLMs in federated settings and the existing universal FL frameworks.
To bridge these gaps, we introduce our open-source \package, \ours, with rich functionalities and extensibilities, which supports federated fine-tuning LLMs under various FL scenarios.
We conduct extensive experiments to demonstrate the utility of our \package and gain insights into how to fine-tune LLMs in FL settings.
Based on the findings from these experimental results, we outline some promising directions for future research in federated LLM fine-tuning to advance the FL and LLM community.
We have released \ours at \href{https://github.com/alibaba/FederatedScope/tree/llm}{https://github.com/alibaba/FederatedScope/tree/llm} for promoting further research.

\newpage
\bibliography{iclr2023_conference}

\begin{thebibliography}{95}
\providecommand{\natexlab}[1]{#1}
\providecommand{\url}[1]{\texttt{#1}}
\expandafter\ifx\csname urlstyle\endcsname\relax
  \providecommand{\doi}[1]{doi: #1}\else
  \providecommand{\doi}{doi: \begingroup \urlstyle{rm}\Url}\fi

\bibitem[Alex et~al.(2021)Alex, Lifland, Tunstall, Thakur, Maham, Riedel, Hine,
  Ashurst, Sedille, Carlier, et~al.]{alex2021raft}
Neel Alex, Eli Lifland, Lewis Tunstall, Abhishek Thakur, Pegah Maham, C~Jess
  Riedel, Emmie Hine, Carolyn Ashurst, Paul Sedille, Alexis Carlier, et~al.
\newblock {RAFT: A Real-World Few-Shot Text Classification Benchmark}.
\newblock \emph{arXiv preprint arXiv:2109.14076}, 2021.

\bibitem[Anthropic(2023)]{claude}
Anthropic.
\newblock {Introducing Claude}, 2023.

\bibitem[Bajaj et~al.(2016)Bajaj, Campos, Craswell, Deng, Gao, Liu, Majumder,
  McNamara, Mitra, Nguyen, et~al.]{Msmarco}
Payal Bajaj, Daniel Campos, Nick Craswell, Li~Deng, Jianfeng Gao, Xiaodong Liu,
  Rangan Majumder, Andrew McNamara, Bhaskar Mitra, Tri Nguyen, et~al.
\newblock {MS MARCO: A Human Generated MAchine Reading COmprehension Dataset}.
\newblock \emph{arXiv preprint arXiv:1611.09268}, 2016.

\bibitem[Bergstra \& Bengio(2012)Bergstra and Bengio]{randomsearch}
James Bergstra and Yoshua Bengio.
\newblock {Random Search for Hyper-Parameter Optimization}.
\newblock \emph{Journal of machine learning research}, 13\penalty0 (2), 2012.

\bibitem[Bi et~al.(2023)Bi, Xie, Zhang, Chen, Gu, and Tian]{pangu}
Kaifeng Bi, Lingxi Xie, Hengheng Zhang, Xin Chen, Xiaotao Gu, and Qi~Tian.
\newblock {Accurate medium-range global weather forecasting with 3D neural
  networks}.
\newblock \emph{Nature}, 619:\penalty0 533 -- 538, 2023.

\bibitem[Bonawitz et~al.(2019)Bonawitz, Eichner, Grieskamp, Huba, Ingerman,
  Ivanov, Kiddon, Kone{\v{c}}n{\`y}, Mazzocchi, McMahan, et~al.]{tff}
Keith Bonawitz, Hubert Eichner, Wolfgang Grieskamp, Dzmitry Huba, Alex
  Ingerman, Vladimir Ivanov, Chloe Kiddon, Jakub Kone{\v{c}}n{\`y}, Stefano
  Mazzocchi, Brendan McMahan, et~al.
\newblock Towards federated learning at scale: System design.
\newblock In \emph{Proc.\ of machine learning and systems (MLSys'19)},
  volume~1, pp.\  374--388, 2019.

\bibitem[Borkan et~al.(2019)Borkan, Dixon, Sorensen, Thain, and
  Vasserman]{civilcomments}
Daniel Borkan, Lucas Dixon, Jeffrey Sorensen, Nithum Thain, and Lucy Vasserman.
\newblock {Nuanced Metrics for Measuring Unintended Bias with Real Data for
  Text Classification}.
\newblock In \emph{Companion Proceedings of the World Wide Web Conference
  (WWW'19)}, pp.\  491--500, 2019.

\bibitem[Brown et~al.(2020)Brown, Mann, Ryder, Subbiah, Kaplan, Dhariwal,
  Neelakantan, Shyam, Sastry, Askell, et~al.]{gpt3}
Tom Brown, Benjamin Mann, Nick Ryder, Melanie Subbiah, Jared~D Kaplan, Prafulla
  Dhariwal, Arvind Neelakantan, Pranav Shyam, Girish Sastry, Amanda Askell,
  et~al.
\newblock {Language Models are Few-Shot Learners}.
\newblock In \emph{Proc.\ of the Advances in Neural Information Processing
  Systems (NeurIPS'20)}, volume~33, pp.\  1877--1901, 2020.

\bibitem[Caldas et~al.(2018)Caldas, Duddu, Wu, Li, Kone{\v{c}}n{\`y}, McMahan,
  Smith, and Talwalkar]{leaf}
Sebastian Caldas, Sai Meher~Karthik Duddu, Peter Wu, Tian Li, Jakub
  Kone{\v{c}}n{\`y}, H~Brendan McMahan, Virginia Smith, and Ameet Talwalkar.
\newblock {Leaf: A benchmark for federated settings}.
\newblock \emph{arXiv preprint arXiv:1812.01097}, 2018.

\bibitem[Chaudhary(2023)]{codealpaca}
Sahil Chaudhary.
\newblock Code alpaca: An instruction-following llama model for code
  generation.
\newblock \url{https://github.com/sahil280114/codealpaca}, 2023.

\bibitem[Chen et~al.(2022)Chen, Gao, Kuang, Li, and Ding]{pFL-bench}
Daoyuan Chen, Dawei Gao, Weirui Kuang, Yaliang Li, and Bolin Ding.
\newblock {pFL-Bench: A Comprehensive Benchmark for Personalized Federated
  Learning}.
\newblock In \emph{Proc.\ of the Advances in Neural Information Processing
  Systems (NeurIPS'22)}, pp.\  9344--9360, 2022.

\bibitem[Chen et~al.(2023)Chen, Gao, Xie, Pan, Li, Li, Ding, and
  Zhou]{chen2023fsreal}
Daoyuan Chen, Dawei Gao, Yuexiang Xie, Xuchen Pan, Zitao Li, Yaliang Li, Bolin
  Ding, and Jingren Zhou.
\newblock {FS-Real: Towards Real-World Cross-Device Federated Learning}.
\newblock In \emph{Proc.\ of the ACM SIGKDD International Conference on
  Knowledge Discovery and Data Mining (KDD'23)}, pp.\  3829–3841, 2023.

\bibitem[Chen et~al.(2021)Chen, Tworek, Jun, et~al.]{HumanEval}
Mark Chen, Jerry Tworek, Heewoo Jun, et~al.
\newblock {Evaluating Large Language Models Trained on Code}.
\newblock \emph{arXiv perprint arXiv:2107.03374}, 2021.

\bibitem[Cheng et~al.(2021)Cheng, Fan, Jin, Liu, Chen, Papadopoulos, and
  Yang]{cheng2021secureboost}
Kewei Cheng, Tao Fan, Yilun Jin, Yang Liu, Tianjian Chen, Dimitrios
  Papadopoulos, and Qiang Yang.
\newblock Secureboost: A lossless federated learning framework.
\newblock \emph{IEEE Intelligent Systems}, 36:\penalty0 87--98, 2021.

\bibitem[Choi et~al.(2018)Choi, He, Iyyer, Yatskar, Yih, Choi, Liang, and
  Zettlemoyer]{choi2018quac}
Eunsol Choi, He~He, Mohit Iyyer, Mark Yatskar, Wen-tau Yih, Yejin Choi, Percy
  Liang, and Luke Zettlemoyer.
\newblock {QuAC: Question Answering in Context}.
\newblock In \emph{Proc.\ of the Conference on Empirical Methods in Natural
  Language Processing (EMNLP'18)}, pp.\  2174--2184, 2018.

\bibitem[Chowdhery et~al.(2022)Chowdhery, Narang, Devlin, et~al.]{palm}
Aakanksha Chowdhery, Sharan Narang, Jacob Devlin, et~al.
\newblock {PaLM: Scaling Language Modeling with Pathways}.
\newblock \emph{arXiv preprint arXiv:2204.02311}, 2022.

\bibitem[Clark et~al.(2019)Clark, Lee, Chang, Kwiatkowski, Collins, and
  Toutanova]{clark2019boolq}
Christopher Clark, Kenton Lee, Ming-Wei Chang, Tom Kwiatkowski, Michael
  Collins, and Kristina Toutanova.
\newblock {BoolQ: Exploring the Surprising Difficulty of Natural Yes/No
  Questions}.
\newblock In \emph{Proc.\ of the Conference of the North {A}merican Chapter of
  the Association for Computational Linguistics: Human Language Technologies
  (NAACL'19)}, pp.\  2924--2936, 2019.

\bibitem[Cobbe et~al.(2021)Cobbe, Kosaraju, Bavarian, Chen, Jun, Kaiser,
  Plappert, Tworek, Hilton, Nakano, Hesse, and Schulman]{cobbe2021gsm8k}
Karl Cobbe, Vineet Kosaraju, Mohammad Bavarian, Mark Chen, Heewoo Jun, Lukasz
  Kaiser, Matthias Plappert, Jerry Tworek, Jacob Hilton, Reiichiro Nakano,
  Christopher Hesse, and John Schulman.
\newblock {Training Verifiers to Solve Math Word Problems}.
\newblock \emph{arXiv preprint arXiv:2110.14168}, 2021.

\bibitem[Computer(2023)]{together2023redpajama}
Together Computer.
\newblock {RedPajama: An Open Source Recipe to Reproduce LLaMA training
  dataset}.
\newblock \url{https://github.com/togethercomputer/RedPajama-Data}, 2023.

\bibitem[Conover et~al.(2023)Conover, Hayes, Mathur, Xie, Wan, Shah, Ghodsi,
  Wendell, Zaharia, and Xin]{DatabricksBlog2023DollyV2}
Mike Conover, Matt Hayes, Ankit Mathur, Jianwei Xie, Jun Wan, Sam Shah, Ali
  Ghodsi, Patrick Wendell, Matei Zaharia, and Reynold Xin.
\newblock {Free Dolly: Introducing the World's First Truly Open
  Instruction-Tuned LLM}, 2023.

\bibitem[Dai et~al.(2020)Dai, Low, and Jaillet]{FTS}
Zhongxiang Dai, Bryan Kian~Hsiang Low, and Patrick Jaillet.
\newblock {Federated Bayesian Optimization via Thompson Sampling}.
\newblock In \emph{Proc.\ of the Advances in Neural Information Processing
  Systems (NeurIPS'20)}, volume~33, pp.\  9687--9699, 2020.

\bibitem[Deutsch(1996{\natexlab{a}})]{deflate}
L.~Peter Deutsch.
\newblock {DEFLATE Compressed Data Format Specification version 1.3}.
\newblock RFC 1951, Network Working Group, 1996{\natexlab{a}}.

\bibitem[Deutsch(1996{\natexlab{b}})]{gzip}
L.~Peter Deutsch.
\newblock {GZIP file format specification version 4.3}.
\newblock RFC 1952, Network Working Group, 1996{\natexlab{b}}.

\bibitem[Devlin et~al.(2019)Devlin, Chang, Lee, and Toutanova]{bert}
Jacob Devlin, Ming-Wei Chang, Kenton Lee, and Kristina Toutanova.
\newblock {BERT: Pre-training of Deep Bidirectional Transformers for Language
  Understanding}.
\newblock In \emph{Proc.\ of NAACL-HLT (NAACL-HLT'19)}, pp.\  4171--4186, 2019.

\bibitem[Dong et~al.(2022)Dong, Xie, Ding, Shen, and Li]{dong2022collaborating}
Chenhe Dong, Yuexiang Xie, Bolin Ding, Ying Shen, and Yaliang Li.
\newblock {Collaborating Heterogeneous Natural Language Processing Tasks via
  Federated Learning}.
\newblock \emph{arXiv preprint arXiv:2212.05789}, 2022.

\bibitem[Driess et~al.(2023)Driess, Xia, Sajjadi, Lynch, Chowdhery, Ichter,
  Wahid, Tompson, Vuong, Yu, et~al.]{mm1}
Danny Driess, Fei Xia, Mehdi~SM Sajjadi, Corey Lynch, Aakanksha Chowdhery,
  Brian Ichter, Ayzaan Wahid, Jonathan Tompson, Quan Vuong, Tianhe Yu, et~al.
\newblock {PaLM-E: An Embodied Multimodal Language Model}.
\newblock \emph{arXiv preprint arXiv:2303.03378}, 2023.

\bibitem[Fang et~al.(2021)Fang, Zhao, Tan, Chen, Yu, Wang, Wang, Zhou, and
  Zhang]{fang2021large}
Wenjing Fang, Derun Zhao, Jin Tan, Chaochao Chen, Chaofan Yu, Li~Wang, Lei
  Wang, Jun Zhou, and Benyu Zhang.
\newblock Large-scale secure xgb for vertical federated learning.
\newblock In \emph{Proc.\ of the Conference on Information \& Knowledge
  Management (CIKM'21)}, pp.\  443--452, 2021.

\bibitem[Gao et~al.(2023)Gao, Chen, Li, Xie, Pan, Li, Ding, and
  Zhou]{gao2023fsreal}
Dawei Gao, Daoyuan Chen, Zitao Li, Yuexiang Xie, Xuchen Pan, Yaliang Li, Bolin
  Ding, and Jingren Zhou.
\newblock {FS-Real: A Real-World Cross-Device Federated Learning Platform}.
\newblock \emph{PVLDB}, 16\penalty0 (12):\penalty0 4046–4049, 2023.

\bibitem[Hendrycks et~al.(2021{\natexlab{a}})Hendrycks, Burns, Basart, Critch,
  Li, Song, and Steinhardt]{hendrycks2021ethics}
Dan Hendrycks, Collin Burns, Steven Basart, Andrew Critch, Jerry Li, Dawn Song,
  and Jacob Steinhardt.
\newblock {Aligning AI With Shared Human Values}.
\newblock In \emph{Proc.\ of the International Conference on Learning
  Representations (ICLR'21)}, 2021{\natexlab{a}}.

\bibitem[Hendrycks et~al.(2021{\natexlab{b}})Hendrycks, Burns, Basart, Zou,
  Mazeika, Song, and Steinhardt]{hendryckstest2021}
Dan Hendrycks, Collin Burns, Steven Basart, Andy Zou, Mantas Mazeika, Dawn
  Song, and Jacob Steinhardt.
\newblock {Measuring Massive Multitask Language Understanding}.
\newblock In \emph{Proc.\ of the International Conference on Learning
  Representations (ICLR'21)}, 2021{\natexlab{b}}.

\bibitem[Hermann et~al.(2015)Hermann, Kocisky, Grefenstette, Espeholt, Kay,
  Suleyman, and Blunsom]{cnndn2}
Karl~Moritz Hermann, Tomas Kocisky, Edward Grefenstette, Lasse Espeholt, Will
  Kay, Mustafa Suleyman, and Phil Blunsom.
\newblock {Teaching Machines to Read and Comprehend}.
\newblock In \emph{Proc.\ of the Advances in Neural Information Processing
  Systems (NeurIPS'15)}, volume~28, pp.\  1693--1701, 2015.

\bibitem[Houlsby et~al.(2019)Houlsby, Giurgiu, Jastrzebski, Morrone,
  De~Laroussilhe, Gesmundo, Attariyan, and Gelly]{adapter}
Neil Houlsby, Andrei Giurgiu, Stanislaw Jastrzebski, Bruna Morrone, Quentin
  De~Laroussilhe, Andrea Gesmundo, Mona Attariyan, and Sylvain Gelly.
\newblock {Parameter-Efficient Transfer Learning for NLP}.
\newblock In \emph{Proc.\ of the International Conference on Machine Learning
  (ICML'21)}, pp.\  2790--2799, 2019.

\bibitem[Hu et~al.(2022)Hu, Wallis, Allen-Zhu, Li, Wang, Wang, Chen,
  et~al.]{lora}
Edward~J Hu, Phillip Wallis, Zeyuan Allen-Zhu, Yuanzhi Li, Shean Wang, Lu~Wang,
  Weizhu Chen, et~al.
\newblock {LoRA: Low-Rank Adaptation of Large Language Models}.
\newblock In \emph{Proc.\ of the International Conference on Learning
  Representations (ICLR'22)}, 2022.

\bibitem[Huang et~al.(2023)Huang, Dong, Wang, Hao, Singhal, Ma, Lv, Cui,
  Mohammed, Liu, et~al.]{mm2}
Shaohan Huang, Li~Dong, Wenhui Wang, Yaru Hao, Saksham Singhal, Shuming Ma,
  Tengchao Lv, Lei Cui, Owais~Khan Mohammed, Qiang Liu, et~al.
\newblock {Language Is Not All You Need: Aligning Perception with Language
  Models}.
\newblock \emph{arXiv preprint arXiv:2302.14045}, 2023.

\bibitem[Husain et~al.(2019)Husain, Wu, Gazit, Allamanis, and
  Brockschmidt]{CodeSearchNet}
Hamel Husain, Ho-Hsiang Wu, Tiferet Gazit, Miltiadis Allamanis, and Marc
  Brockschmidt.
\newblock {CodeSearchNet Challenge: Evaluating the State of Semantic Code
  Search}.
\newblock \emph{arXiv preprint arXiv:1909.09436}, 2019.

\bibitem[Ippolito et~al.(2022)Ippolito, Yuan, Coenen, and
  Burnam]{CreativeWriting}
Daphne Ippolito, Ann Yuan, Andy Coenen, and Sehmon Burnam.
\newblock {Creative Writing with an AI-Powered Writing Assistant: Perspectives
  from Professional Writers}.
\newblock \emph{arXiv preprint arXiv:2211.05030}, 2022.

\bibitem[Jamieson \& Talwalkar(2016)Jamieson and
  Talwalkar]{successivehalvingalgorithm}
Kevin Jamieson and Ameet Talwalkar.
\newblock {Non-stochastic Best Arm Identification and Hyperparameter
  Optimization}.
\newblock In \emph{Proc.\ of the Artificial intelligence and statistics
  (AISTATS'16)}, pp.\  240--248, 2016.

\bibitem[Ji et~al.(2023)Ji, Lee, Frieske, Yu, Su, Xu, Ishii, Bang, Madotto, and
  Fung]{hallucination}
Ziwei Ji, Nayeon Lee, Rita Frieske, Tiezheng Yu, Dan Su, Yan Xu, Etsuko Ishii,
  Ye~Jin Bang, Andrea Madotto, and Pascale Fung.
\newblock {Survey of Hallucination in Natural Language Generation}.
\newblock \emph{{ACM} Computing Surveys}, 55:\penalty0 1--38, 2023.

\bibitem[Karimi~Mahabadi et~al.(2021)Karimi~Mahabadi, Henderson, and
  Ruder]{compacter}
Rabeeh Karimi~Mahabadi, James Henderson, and Sebastian Ruder.
\newblock {Compacter: Efficient Low-Rank Hypercomplex Adapter Layers}.
\newblock In \emph{Proc.\ of the Advances in Neural Information Processing
  Systems (NeurIPS'21)}, volume~34, pp.\  1022--1035, 2021.

\bibitem[Ko{\v{c}}isk{\`y} et~al.(2018)Ko{\v{c}}isk{\`y}, Schwarz, Blunsom,
  Dyer, Hermann, Melis, and Grefenstette]{narrativeqa}
Tom{\'a}{\v{s}} Ko{\v{c}}isk{\`y}, Jonathan Schwarz, Phil Blunsom, Chris Dyer,
  Karl~Moritz Hermann, G{\'a}bor Melis, and Edward Grefenstette.
\newblock {The NarrativeQA Reading Comprehension Challenge}.
\newblock \emph{Transactions of the Association for Computational Linguistics},
  6:\penalty0 317--328, 2018.

\bibitem[Kone{\v{c}}n{\'y} et~al.(2016)Kone{\v{c}}n{\'y}, McMahan, Ramage, and
  Richt{\'{a}}rik]{fl1}
Jakub Kone{\v{c}}n{\'y}, H.~Brendan McMahan, Daniel Ramage, and Peter
  Richt{\'{a}}rik.
\newblock {Federated Optimization: Distributed Machine Learning for On-Device
  Intelligence}.
\newblock \emph{arXiv preprint arXiv:1610.02527}, 2016.

\bibitem[Kwiatkowski et~al.(2019)Kwiatkowski, Palomaki, Redfield, Collins,
  Parikh, Alberti, Epstein, Polosukhin, Devlin, Lee,
  et~al.]{kwiatkowski2019natural}
Tom Kwiatkowski, Jennimaria Palomaki, Olivia Redfield, Michael Collins, Ankur
  Parikh, Chris Alberti, Danielle Epstein, Illia Polosukhin, Jacob Devlin,
  Kenton Lee, et~al.
\newblock {Natural Questions: A Benchmark for Question Answering Research}.
\newblock \emph{Transactions of the Association for Computational Linguistics},
  7:\penalty0 453--466, 2019.

\bibitem[Lai et~al.(2022)Lai, Dai, Singapuram, Liu, Zhu, Madhyastha, and
  Chowdhury]{Lai2021FedScaleBM}
Fan Lai, Yinwei Dai, Sanjay Singapuram, Jiachen Liu, Xiangfeng Zhu, Harsha
  Madhyastha, and Mosharaf Chowdhury.
\newblock {FedScale: Benchmarking Model and System Performance of Federated
  Learning at Scale}.
\newblock In \emph{Proc.\ of the International Conference on Machine Learning
  (ICML'22)}, volume 162, pp.\  11814--11827, 2022.

\bibitem[Lee et~al.(2022)Lee, Liang, and Yang]{CoAuthor}
Mina Lee, Percy Liang, and Qian Yang.
\newblock {C}o{A}uthor: {Designing a Human-AI Collaborative Writing Dataset for
  Exploring Language Model Capabilities}.
\newblock In \emph{Proc.\ of the CHI Conference on Human Factors in Computing
  Systems (CHI'22)}, 2022.

\bibitem[Lester et~al.(2021)Lester, Al-Rfou, and Constant]{prompttuning}
Brian Lester, Rami Al-Rfou, and Noah Constant.
\newblock {The Power of Scale for Parameter-Efficient Prompt Tuning}.
\newblock In \emph{Proc.\ of the Conference on Empirical Methods in Natural
  Language Processing (EMNLP'21)}, pp.\  3045--3059, 2021.

\bibitem[Li et~al.(2021)Li, Hu, Beirami, and Smith]{ditto}
Tian Li, Shengyuan Hu, Ahmad Beirami, and Virginia Smith.
\newblock {Ditto: Fair and Robust Federated Learning Through Personalization}.
\newblock In \emph{Proc.\ of the International Conference on Machine Learning
  (ICML'21)}, volume 139, pp.\  6357--6368, 2021.

\bibitem[Li \& Liang(2021)Li and Liang]{PrefixTuning}
Xiang~Lisa Li and Percy Liang.
\newblock {Prefix-Tuning: Optimizing Continuous Prompts for Generation}.
\newblock In \emph{Proc.\ of the Annual Meeting of the Association for
  Computational Linguistics and the International Joint Conference on Natural
  Language Processing (ACL-IJCNLP'21)}, pp.\  4582--4597, 2021.

\bibitem[Liang et~al.(2022)Liang, Bommasani, Lee, Tsipras, Soylu, Yasunaga,
  Zhang, Narayanan, Wu, Kumar, et~al.]{helm}
Percy Liang, Rishi Bommasani, Tony Lee, Dimitris Tsipras, Dilara Soylu,
  Michihiro Yasunaga, Yian Zhang, Deepak Narayanan, Yuhuai Wu, Ananya Kumar,
  et~al.
\newblock {Holistic Evaluation of Language Models}.
\newblock \emph{arXiv preprint arXiv:2211.09110}, 2022.

\bibitem[Lin et~al.(2021)Lin, Hilton, and Evans]{lin2021truthfulqa}
Stephanie Lin, Jacob Hilton, and Owain Evans.
\newblock {TruthfulQA: Measuring How Models Mimic Human Falsehoods}.
\newblock \emph{arXiv preprint arXiv:2109.07958}, 2021.

\bibitem[Liu et~al.(2021{\natexlab{a}})Liu, Ji, Fu, Du, Yang, and
  Tang]{PTuningV2}
Xiao Liu, Kaixuan Ji, Yicheng Fu, Zhengxiao Du, Zhilin Yang, and Jie Tang.
\newblock {P-Tuning v2: Prompt Tuning Can Be Comparable to Fine-tuning
  Universally Across Scales and Tasks}.
\newblock \emph{arXiv preprint arXiv:2110.07602}, 2021{\natexlab{a}}.

\bibitem[Liu et~al.(2021{\natexlab{b}})Liu, Zheng, Du, Ding, Qian, Yang, and
  Tang]{ptuning}
Xiao Liu, Yanan Zheng, Zhengxiao Du, Ming Ding, Yujie Qian, Zhilin Yang, and
  Jie Tang.
\newblock {GPT Understands, Too}.
\newblock \emph{arXiv preprint arXiv:2103.10385}, 2021{\natexlab{b}}.

\bibitem[Liu et~al.(2019)Liu, Ott, Goyal, Du, Joshi, Chen, Levy, Lewis,
  Zettlemoyer, and Stoyanov]{RoBERTa}
Yinhan Liu, Myle Ott, Naman Goyal, Jingfei Du, Mandar Joshi, Danqi Chen, Omer
  Levy, Mike Lewis, Luke Zettlemoyer, and Veselin Stoyanov.
\newblock {RoBERTa: A Robustly Optimized BERT Pretraining Approach}.
\newblock \emph{arXiv preprint arXiv:1907.11692}, 2019.

\bibitem[Maas et~al.(2011)Maas, Daly, Pham, Huang, Ng, and Potts]{imdb}
Andrew~L. Maas, Raymond~E. Daly, Peter~T. Pham, Dan Huang, Andrew~Y. Ng, and
  Christopher Potts.
\newblock {Learning Word Vectors for Sentiment Analysis}.
\newblock In \emph{Proc.\ of the Annual Meeting of the Association for
  Computational Linguistics: Human Language Technologies (AMACL'11)}, pp.\
  142--150, 2011.

\bibitem[Mangrulkar et~al.(2022)Mangrulkar, Gugger, Debut, Belkada, and
  Paul]{peft}
Sourab Mangrulkar, Sylvain Gugger, Lysandre Debut, Younes Belkada, and Sayak
  Paul.
\newblock {PEFT: State-of-the-art Parameter-Efficient Fine-Tuning methods}.
\newblock \url{https://github.com/huggingface/peft}, 2022.

\bibitem[McMahan et~al.(2017)McMahan, Moore, Ramage, Hampson, and Arcas]{fl2}
H.~Brendan McMahan, Eider Moore, Daniel Ramage, Seth Hampson, and Blaise
  Ag{\"{u}}era~y Arcas.
\newblock {Communication-Efficient Learning of Deep Networks from Decentralized
  Data}.
\newblock In \emph{Proc.\ of the Artificial intelligence and statistics
  (AISTATS'17)}, pp.\  1273--1282, 2017.

\bibitem[Micikevicius et~al.(2018)Micikevicius, Narang, Alben, Diamos, Elsen,
  Garcia, Ginsburg, Houston, Kuchaiev, Venkatesh, et~al.]{mixprec}
Paulius Micikevicius, Sharan Narang, Jonah Alben, Gregory Diamos, Erich Elsen,
  David Garcia, Boris Ginsburg, Michael Houston, Oleksii Kuchaiev, Ganesh
  Venkatesh, et~al.
\newblock {Mixed Precision Training}.
\newblock In \emph{Proc.\ of the International Conference on Learning
  Representations (ICLR'18)}, 2018.

\bibitem[Mihaylov et~al.(2018)Mihaylov, Clark, Khot, and Sabharwal]{openbookqa}
Todor Mihaylov, Peter Clark, Tushar Khot, and Ashish Sabharwal.
\newblock {Can a Suit of Armor Conduct Electricity? A New Dataset for Open Book
  Question Answering}.
\newblock \emph{arXiv preprint arXiv:1809.02789}, 2018.

\bibitem[Narayan et~al.(2018)Narayan, Cohen, and Lapata]{xsum-emnlp}
Shashi Narayan, Shay~B. Cohen, and Mirella Lapata.
\newblock {Don't Give Me the Details, Just the Summary! {T}opic-Aware
  Convolutional Neural Networks for Extreme Summarization}.
\newblock In \emph{Proc.\ of the Conference on Empirical Methods in Natural
  Language Processing (EMNLP'18)}, pp.\  1797--1807, 2018.

\bibitem[OpenAI(2022)]{gpt35}
OpenAI.
\newblock {Introducing ChatGPT}, 2022.

\bibitem[OpenAI(2023)]{gpt4}
OpenAI.
\newblock {GPT-4 Technical Report}.
\newblock \emph{arXiv preprint arXiv:2303.08774}, 2023.

\bibitem[Paszke et~al.(2019)Paszke, Gross, Massa, Lerer, Bradbury, Chanan,
  Killeen, Lin, Gimelshein, Antiga, et~al.]{pytorch}
Adam Paszke, Sam Gross, Francisco Massa, Adam Lerer, James Bradbury, Gregory
  Chanan, Trevor Killeen, Zeming Lin, Natalia Gimelshein, Luca Antiga, et~al.
\newblock {PyTorch: An Imperative Style, High-Performance Deep Learning
  Library}.
\newblock In \emph{Proc.\ of the Advances in Neural Information Processing
  Systems (NeurIPS'19)}, pp.\  8024--8035, 2019.

\bibitem[Patil et~al.(2023)Patil, Zhang, Wang, and Gonzalez]{Gorilla}
Shishir~G. Patil, Tianjun Zhang, Xin Wang, and Joseph~E. Gonzalez.
\newblock {Gorilla: Large Language Model Connected with Massive APIs}.
\newblock \emph{arXiv preprint arXiv:2305.15334}, 2023.

\bibitem[Peters et~al.(2018)Peters, Neumann, Iyyer, Gardner, Clark, Lee, and
  Zettlemoyer]{ELMo}
Matthew~E. Peters, Mark Neumann, Mohit Iyyer, Matt Gardner, Christopher Clark,
  Kenton Lee, and Luke Zettlemoyer.
\newblock {Deep Contextualized Word Representations}.
\newblock In \emph{Proc.\ of North American Chapter of the Association for
  Computational Linguistics (NAACL-HLT'18)}, pp.\  2227--2237, 2018.

\bibitem[Pfeiffer et~al.(2020{\natexlab{a}})Pfeiffer, Kamath, R{\"u}ckl{\'e},
  Cho, and Gurevych]{fusion}
Jonas Pfeiffer, Aishwarya Kamath, Andreas R{\"u}ckl{\'e}, Kyunghyun Cho, and
  Iryna Gurevych.
\newblock {AdapterFusion: Non-Destructive Task Composition for Transfer
  Learning}.
\newblock \emph{arXiv preprint arXiv:2005.00247}, 2020{\natexlab{a}}.

\bibitem[Pfeiffer et~al.(2020{\natexlab{b}})Pfeiffer, R\"uckl\'{e}, Poth,
  Kamath, Vuli\'{c}, Ruder, Cho, and Gurevych]{adapterhub}
Jonas Pfeiffer, Andreas R\"uckl\'{e}, Clifton Poth, Aishwarya Kamath, Ivan
  Vuli\'{c}, Sebastian Ruder, Kyunghyun Cho, and Iryna Gurevych.
\newblock {AdapterHub: A Framework for Adapting Transformers}.
\newblock In \emph{Proceedings of the Conference on Empirical Methods in
  Natural Language Processing: Systems Demonstrations (EMNLP'20)}, pp.\
  46--54, 2020{\natexlab{b}}.

\bibitem[Pfeiffer et~al.(2020{\natexlab{c}})Pfeiffer, Vuli{\'c}, Gurevych, and
  Ruder]{invertible}
Jonas Pfeiffer, Ivan Vuli{\'c}, Iryna Gurevych, and Sebastian Ruder.
\newblock {MAD-X}: {A}n {A}dapter-{B}ased {F}ramework for {M}ulti-{T}ask
  {C}ross-{L}ingual {T}ransfer.
\newblock In \emph{Proc.\ of the Conference on Empirical Methods in Natural
  Language Processing (EMNLP'20)}, pp.\  7654--7673, 2020{\natexlab{c}}.

\bibitem[Qin et~al.(2023)Qin, Liang, Ye, Zhu, Yan, Lu, Lin, Cong, Tang, Qian,
  et~al.]{ToolLLM}
Yujia Qin, Shihao Liang, Yining Ye, Kunlun Zhu, Lan Yan, Yaxi Lu, Yankai Lin,
  Xin Cong, Xiangru Tang, Bill Qian, et~al.
\newblock {T}ool{LLM}: {Facilitating Large Language Models to Master 16000+
  Real-world APIs}.
\newblock \emph{arXiv preprint arXiv:2307.16789}, 2023.

\bibitem[Radford \& Narasimhan(2018)Radford and Narasimhan]{GPT}
Alec Radford and Karthik Narasimhan.
\newblock {Improving Language Understanding by Generative Pre-Training}.
\newblock 2018.

\bibitem[Radford et~al.(2019)Radford, Wu, Child, Luan, Amodei, and
  Sutskever]{GPT2}
Alec Radford, Jeff Wu, Rewon Child, David Luan, Dario Amodei, and Ilya
  Sutskever.
\newblock {Language Models are Unsupervised Multitask Learners}.
\newblock 2019.

\bibitem[Rasley et~al.(2020)Rasley, Rajbhandari, Ruwase, and He]{deepspeed}
Jeff Rasley, Samyam Rajbhandari, Olatunji Ruwase, and Yuxiong He.
\newblock {DeepSpeed: System Optimizations Enable Training Deep Learning Models
  with Over 100 Billion Parameters}.
\newblock In \emph{Proc.\ of the ACM SIGKDD International Conference on
  Knowledge Discovery and Data Mining (KDD'20)}, pp.\  3505–3506, 2020.

\bibitem[Ren et~al.(2021)Ren, Rajbhandari, Aminabadi, Ruwase, Yang, Zhang, Li,
  and He]{offloading}
Jie Ren, Samyam Rajbhandari, Reza~Yazdani Aminabadi, Olatunji Ruwase,
  Shuangyang Yang, Minjia Zhang, Dong Li, and Yuxiong He.
\newblock {ZeRO-Offload: Democratizing Billion-Scale Model Training}.
\newblock In \emph{Proc.\ of the USENIX Annual Technical Conference (USENIX
  ATC'21)}, pp.\  551--564, 2021.

\bibitem[Ruebsamen(2023)]{cleanedalpaca}
Gene Ruebsamen.
\newblock {Cleaned Alpaca Dataset}.
\newblock \url{https://github.com/gururise/AlpacaDataCleaned}, 2023.

\bibitem[Ryffel et~al.(2018)Ryffel, Trask, Dahl, Wagner, Mancuso, Rueckert, and
  Passerat-Palmbach]{pysyft}
Th{\'e}o Ryffel, Andrew Trask, Morten Dahl, Bobby Wagner, Jason~V. Mancuso,
  Daniel Rueckert, and Jonathan Passerat-Palmbach.
\newblock A generic framework for privacy preserving deep learning.
\newblock \emph{arXiv preprint arXiv:1811.04017}, 2018.

\bibitem[Scao et~al.(2022)Scao, Fan, Akiki, et~al.]{bloom}
Teven~Le Scao, Angela Fan, Christopher Akiki, et~al.
\newblock {BLOOM: A 176B-Parameter Open-Access Multilingual Language Model}.
\newblock \emph{arXiv preprint arXiv:2211.05100}, 2022.

\bibitem[See et~al.(2017)See, Liu, and Manning]{cnndn1}
Abigail See, Peter~J. Liu, and Christopher~D. Manning.
\newblock {Get To The Point: Summarization with Pointer-Generator Networks}.
\newblock In \emph{Proc.\ of the Annual Meeting of the Association for
  Computational Linguistics (AMACL'17)}, pp.\  1073--1083, 2017.

\bibitem[Shahriari et~al.(2015)Shahriari, Swersky, Wang, Adams, and
  De~Freitas]{bayesianoptimization}
Bobak Shahriari, Kevin Swersky, Ziyu Wang, Ryan~P Adams, and Nando De~Freitas.
\newblock {Taking the Human Out of the Loop: A Review of Bayesian
  Optimization}.
\newblock \emph{Proceedings of the IEEE}, 104\penalty0 (1):\penalty0 148--175,
  2015.

\bibitem[T~Dinh et~al.(2020)T~Dinh, Tran, and Nguyen]{pFedME}
Canh T~Dinh, Nguyen Tran, and Josh Nguyen.
\newblock {Personalized Federated Learning with Moreau Envelopes}.
\newblock In \emph{Proc.\ of the Advances in Neural Information Processing
  Systems (NeurIPS'20)}, volume~33, pp.\  21394--21405, 2020.

\bibitem[Tan et~al.(2022)Tan, Yu, Cui, and Yang]{pFL}
Alysa~Ziying Tan, Han Yu, Lizhen Cui, and Qiang Yang.
\newblock {Towards Personalized Federated Learning}.
\newblock \emph{{IEEE} Transactions on Neural Networks and Learning Systems},
  pp.\  1--17, 2022.

\bibitem[Taori et~al.(2023)Taori, Gulrajani, Zhang, Dubois, Li, Guestrin,
  Liang, and Hashimoto]{alpaca}
Rohan Taori, Ishaan Gulrajani, Tianyi Zhang, Yann Dubois, Xuechen Li, Carlos
  Guestrin, Percy Liang, and Tatsunori~B. Hashimoto.
\newblock {Stanford Alpaca: An Instruction-following LLaMA model}.
\newblock \url{https://github.com/tatsu-lab/stanford\_alpaca}, 2023.

\bibitem[Touvron et~al.(2023)Touvron, Lavril, Izacard, Martinet, Lachaux,
  Lacroix, Rozi{\`e}re, Goyal, Hambro, Azhar, et~al.]{llama}
Hugo Touvron, Thibaut Lavril, Gautier Izacard, Xavier Martinet, Marie-Anne
  Lachaux, Timoth{\'e}e Lacroix, Baptiste Rozi{\`e}re, Naman Goyal, Eric
  Hambro, Faisal Azhar, et~al.
\newblock {LLaMA: Open and Efficient Foundation Language Models}.
\newblock \emph{arXiv preprint arXiv:2302.13971}, 2023.

\bibitem[Wang et~al.(2022)Wang, Kuang, Xie, Yao, Li, Ding, and Zhou]{FS-GNN}
Zhen Wang, Weirui Kuang, Yuexiang Xie, Liuyi Yao, Yaliang Li, Bolin Ding, and
  Jingren Zhou.
\newblock {FederatedScope-GNN: Towards a Unified, Comprehensive and Efficient
  Package for Federated Graph Learning}.
\newblock In \emph{Proc.\ of the ACM SIGKDD International Conference on
  Knowledge Discovery and Data Mining (KDD'22)}, 2022.

\bibitem[Wang et~al.(2023)Wang, Kuang, Zhang, Ding, and Li]{FedHPOBench}
Zhen Wang, Weirui Kuang, Ce~Zhang, Bolin Ding, and Yaliang Li.
\newblock {FedHPO-Bench: A Benchmark Suite for Federated Hyperparameter
  Optimization}.
\newblock In \emph{Proc.\ of the International Conference on Machine Learning
  (ICML'23)}, pp.\  35908--35948, 2023.

\bibitem[Wei et~al.(2022)Wei, Wang, Schuurmans, Bosma, Xia, Chi, Le, Zhou,
  et~al.]{wei2022chain}
Jason Wei, Xuezhi Wang, Dale Schuurmans, Maarten Bosma, Fei Xia, Ed~Chi, Quoc~V
  Le, Denny Zhou, et~al.
\newblock {Chain-of-Thought Prompting Elicits Reasoning in Large Language
  Models}.
\newblock In \emph{Proc.\ of the Advances in Neural Information Processing
  Systems (NeurIPS'22)}, volume~35, pp.\  24824--24837, 2022.

\bibitem[Wolf et~al.(2020)Wolf, Debut, Sanh, Chaumond, Delangue, Moi, Cistac,
  Rault, Louf, Funtowicz, Davison, Shleifer, von Platen, Ma, Jernite, Plu, Xu,
  Scao, Gugger, Drame, Lhoest, and Rush]{hftransformers}
Thomas Wolf, Lysandre Debut, Victor Sanh, Julien Chaumond, Clement Delangue,
  Anthony Moi, Pierric Cistac, Tim Rault, Rémi Louf, Morgan Funtowicz, Joe
  Davison, Sam Shleifer, Patrick von Platen, Clara Ma, Yacine Jernite, Julien
  Plu, Canwen Xu, Teven~Le Scao, Sylvain Gugger, Mariama Drame, Quentin Lhoest,
  and Alexander~M. Rush.
\newblock {Transformers: State-of-the-Art Natural Language Processing}.
\newblock In \emph{Proc.\ of the Conference on Empirical Methods in Natural
  Language Processing (EMNLP'20)}, pp.\  38--45, 2020.

\bibitem[Wu et~al.(2023)Wu, Yin, Qi, Wang, Tang, and Duan]{mm3}
Chenfei Wu, Sheng-Kai Yin, Weizhen Qi, Xiaodong Wang, Zecheng Tang, and Nan
  Duan.
\newblock {Visual ChatGPT: Talking, Drawing and Editing with Visual Foundation
  Models}.
\newblock \emph{arXiv preprint arXiv:2303.04671}, 2023.

\bibitem[Wu et~al.(2020)Wu, Cai, Xiao, Chen, and Ooi]{wu13privacy}
Yuncheng Wu, Shaofeng Cai, Xiaokui Xiao, Gang Chen, and Beng~Chin Ooi.
\newblock Privacy preserving vertical federated learning for tree-based models.
\newblock \emph{PVLDB}, 13:\penalty0 2090–2103, 2020.

\bibitem[Xiao et~al.(2023)Xiao, Lin, and Han]{offsite}
Guangxuan Xiao, Ji~Lin, and Song Han.
\newblock {Offsite-Tuning: Transfer Learning without Full Model}.
\newblock \emph{arXiv preprint arXiv:2302.04870}, 2023.

\bibitem[Xie et~al.(2023)Xie, Wang, Chen, Gao, Yao, Kuang, Li, Ding, and
  Zhou]{federatedscope}
Yuexiang Xie, Zhen Wang, Daoyuan Chen, Dawei Gao, Liuyi Yao, Weirui Kuang,
  Yaliang Li, Bolin Ding, and Jingren Zhou.
\newblock {FederatedScope: A Flexible Federated Learning Platform for
  Heterogeneity}.
\newblock \emph{PVLDB}, 16\penalty0 (5):\penalty0 1059–1072, 2023.

\bibitem[Yang et~al.(2019)Yang, Liu, Cheng, Kang, Chen, and Yu]{fl3}
Qiang Yang, Yang Liu, Yong Cheng, Yan Kang, Tianjian Chen, and Han Yu.
\newblock {Federated Learning}.
\newblock \emph{Synthesis Lectures on Artificial Intelligence and Machine
  Learning}, 13:\penalty0 1--207, 2019.

\bibitem[Zellers et~al.(2019)Zellers, Holtzman, Bisk, Farhadi, and
  Choi]{zellers2019hellaswag}
Rowan Zellers, Ari Holtzman, Yonatan Bisk, Ali Farhadi, and Yejin Choi.
\newblock {HellaSwag: Can a Machine Really Finish Your Sentence?}
\newblock In \emph{Proc.\ of the Annual Meeting of the Association for
  Computational Linguistics (AMACL'19)}, 2019.

\bibitem[Zeng et~al.(2023)Zeng, Liu, Du, Wang, Lai, Ding, Yang, Xu, Zheng, Xia,
  et~al.]{GLM}
Aohan Zeng, Xiao Liu, Zhengxiao Du, Zihan Wang, Hanyu Lai, Ming Ding, Zhuoyi
  Yang, Yifan Xu, Wendi Zheng, Xiao Xia, et~al.
\newblock {GLM-130B: An Open Bilingual Pre-trained Model}.
\newblock In \emph{Proc.\ of the International Conference on Learning
  Representations (ICLR'23)}, 2023.

\bibitem[Zhang et~al.(2022)Zhang, Roller, Goyal, Artetxe, Chen, Chen, Dewan,
  Diab, Li, Lin, et~al.]{opt}
Susan Zhang, Stephen Roller, Naman Goyal, Mikel Artetxe, Moya Chen, Shuohui
  Chen, Christopher Dewan, Mona Diab, Xian Li, Xi~Victoria Lin, et~al.
\newblock {OPT: Open Pre-trained Transformer Language Models}.
\newblock \emph{arXiv preprint arXiv:2205.01068}, 2022.

\bibitem[Zhang et~al.(2019)Zhang, Liu, Zhang, Liu, Huang, Zhou, Guo, Kang, Guo,
  Du, and Chen]{quant}
Xishan Zhang, Shaoli Liu, Rui Zhang, Chang Liu, Di~Huang, Shiyi Zhou, Jiaming
  Guo, Yu~Kang, Qi~Guo, Zidong Du, and Yunji Chen.
\newblock {Adaptive Precision Training: Quantify Back Propagation in Neural
  Networks with Fixed-point Numbers}.
\newblock \emph{arXiv preprint arXiv:1911.00361}, 2019.

\bibitem[Zhao et~al.(2023)Zhao, Zhou, Li, Tang, Wang, Hou, Min, Zhang, Zhang,
  Dong, et~al.]{zhao2023survey}
Wayne~Xin Zhao, Kun Zhou, Junyi Li, Tianyi Tang, Xiaolei Wang, Yupeng Hou,
  Yingqian Min, Beichen Zhang, Junjie Zhang, Zican Dong, et~al.
\newblock {A Survey of Large Language Models}.
\newblock \emph{arXiv preprint arXiv:2303.18223}, 2023.

\bibitem[Zhou et~al.(2021)Zhou, Ram, Salonidis, Angel, Samulowitz, and
  Ludwig]{flora}
Yi~Zhou, Parikshit Ram, Theodoros Salonidis, Nathalie~Baracaldo Angel, Horst
  Samulowitz, and Heiko Ludwig.
\newblock {FLoRA: Single-shot Hyper-parameter Optimization for Federated
  Learning}.
\newblock In \emph{Proc.\ of the Advances in Neural Information Processing
  Systems (NeurIPS'21)}, 2021.

\end{thebibliography}
\bibliographystyle{iclr2023_conference}

\newpage
\appendix
\section{Appendix}

\subsection{Fine-tuning Dataset description}
\label{app:data}
In this section, we describe the fine-tuning datasets curated in \ours, and summarize the statistics and information in Table~\ref{tab:dataset}. 
These datasets are derived from existing and widely used fine-tuning datasets that cover diverse domains, such as code, natural language, dialogues, and math problems. 
The curated fine-tuning datasets exhibit different degrees of heterogeneity across clients, which pose various challenges and opportunities for fine-tuning LLMs in FL. 
We describe the construction and characteristics of each dataset in detail below and illustrate their scenarios. 
We will constantly adopt new datasets for fine-tuning LLMs in \ours.

\begin{table*}[htbp]
\renewcommand{\arraystretch}{1.5}
\centering
\caption{Statistics and information of the fine-tuning datasets.}
\resizebox{\textwidth}{!}{
\begin{tabular}{cccccc}
\toprule
Name & \#Client & \#Sample & Split & Domain & Evaluation Dataset\\ \hline
\textit{Fed\databar CodeAlpaca} & 9 & $\sim$8.0k & Meta & Code Generation & \textit{HumanEval} \\
\textit{Fed\databar Dolly} & 8 & $\sim$15.0k & Meta & Generic Language & \textit{HELM}\\ 
\textit{Fed\databar GSM8K-3} & 3 & $\sim$7.5k & IID & CoT & \textit{GSM8K-test} \\ 
\textit{Fed\databar CodeSearchNet} & 6 & $\sim$1880.8k & Meta & Code Generation & \textit{HumanEval} \\ \hline
\textit{Alpaca} & - & $\sim$52.0k & - & Generic Language & \textit{HELM} \\ 
\textit{CleanedAlpaca} & - & $\sim$51.8k & - & Generic Language & \textit{HELM}\\
\bottomrule
\end{tabular}
}
\label{tab:dataset}
\end{table*}

\subsubsection{Federated fine-tuning dataset}
The federated fine-tuning datasets are a collection of curated datasets that we adopt based on the meta-information or some distribution of the original corpora. 
Users can directly use them for federated fine-tuning LLMs.

\noindent\textbf{\textit{Fed\databar CodeAlpaca}} is a federated version of \textit{CodeAlpaca}~\citep{codealpaca}, a code dataset that contains ten programming languages, including C, C\#, C++, Go, Java, PHP, Pascal, Python, Scale, and X86-64 Assemble. We exclude the X86-64 Assembly samples, as they are very scarce in the original corpora. Then, we split the remaining samples into nine subsets based on the language category and assign each subset to one client.

\noindent\textbf{\textit{Fed\databar Dolly}} is a federated corpus dataset derived from \textit{Databricks-dolly-15k}~\citep{DatabricksBlog2023DollyV2}, which comprises eight categories of NLP tasks: brainstorming, classification, closed QA, creative writing, general QA, information extraction, open QA, and summarization. The corpora within each client only belong to one category.

\noindent\textbf{\textit{Fed\databar GSM8K-3}} is built from \textit{GSM8K}~\citep{cobbe2021gsm8k}, which is a mathematical fine-tuning dataset consisting of 7.5K training problems and 1K test problems. We split the training problems into three subsets by the uniform splitter, and assign each subset to one client.

In addition to the three datasets introduced in the main text, we also curate another code dataset, \textit{Fed\databar CodeSearchNet}, as an alternative to \textit{Fed\databar CodeAlpaca}.

\noindent\textbf{\textit{Fed\databar CodeSearchNet}} is built from \textit{CodeSearchNet}~\citep{CodeSearchNet}, which is a large-scale code dataset of functions and their associated documentation for six programming languages: Go, Java, JavaScript, PHP, Python, and Ruby. The data are extracted from open-source projects on GitHub. 
Similar to \textit{Fed\databar CodeAlpaca}, we split the samples into six subsets according to the language category and allocate each subset to one client.

\subsubsection{Centralized fine-tuning dataset}
The centralized fine-tuning datasets are a collection of corpora that we have collected from the Internet without any prior partitioning. 
Users can use our provided splitters to customize the data partition according to different criteria, such as heterogeneity, number balance, root verb, etc.

\noindent\textbf{\textit{Alpaca}}~\citep{alpaca} is a fine-tuning dataset containing natural language questions and responses for various NLP tasks such as text generation, translation, and open QA. The dataset covers a wide range of domains, such as math, text processing, code generation, etc.

\noindent\textbf{\textit{CleanedAlpaca}}~\citep{cleanedalpaca} is a fine-tuning dataset that improves the quality and usefulness of the original \textit{Alpaca} dataset. It is expected to be more reliable and consistent for fine-tuning LLMs.

\subsection{Evaluation task description}
\label{app:eval}
We believe that fine-tuning LLMs should either improve their generic language capabilities or improve their domain-specific capabilities for one particular downstream task.
Thus, we use three different evaluation tasks for benchmarking fine-tuned LLMs. 

\noindent\textbf{Evaluation task for code generation capability.} \textit{HumanEval}~\citep{HumanEval} is to measure whether the code generated by LLMs is correct or not. 
It contains an evaluation dataset and an accordingly metric Pass@k score for assessing the performance of LLMs on code generation capability. 
Specifically, the model generates $m$ samples per task and denotes $c$ as the number of correct samples generated by LLM. Then
$$
\text{Pass@} \mathrm{k}:=\mathbb{E}_{\text {Problems}}\left[1-\frac{\left(\begin{array}{c}
m-c \\
k
\end{array}\right)}{\left(\begin{array}{c}
m \\
k
\end{array}\right)}\right] \text {.}
$$
In practice, we set $m=5$ and use the Pass@1 score as the evaluation score, i.e.,
$$
\text{Pass@1}:=\mathbb{E}_{\text {Problems}}\left[\frac{c}{5}\right] \text {.}
$$

\noindent\textbf{Evaluation task for generic language capability.} We adapt \textit{HELM}~\citep{helm}, including 16 subtasks, in \ours to evaluate the generic language capability of fine-tuned LLMs. 
To be more precise, the subtasks we use are \textit{MMLU}~\citep{hendryckstest2021, hendrycks2021ethics}, \textit{BoolQ}~\citep{clark2019boolq}, \textit{NarrativeQA}~\citep{narrativeqa}, \textit{NaturalQuestions (closed-book)}~\citep{kwiatkowski2019natural} , \textit{NaturalQuestions (open-book)}~\citep{kwiatkowski2019natural} , \textit{QuAC}~\citep{choi2018quac}, \textit{HellaSwag}~\citep{zellers2019hellaswag}, \textit{OpenbookQA}~\citep{openbookqa}, \textit{TruthfulQA}~\citep{lin2021truthfulqa}, \textit{MS MARCO (regular)}~\citep{Msmarco}, \textit{MS MARCO (TREC)}~\citep{Msmarco}, \textit{CNN/DailyMail}~\citep{cnndn1, cnndn2}, \textit{XSUM}~\citep{xsum-emnlp}, \textit{IMDB}~\citep{imdb}, \textit{CivilComments}~\citep{civilcomments}, and \textit{RAFT}~\citep{alex2021raft}. For each task, we randomly use 100 samples for evaluation. The evaluation score for \textit{HELM} is a mixture of metric scores on these 16 subtasks. Specifically, in \textit{HELM}, \textit{MMLU}, \textit{BoolQ}, \textit{HellaSwag}, \textit{OpenbookQA}, \textit{TruthfulQA}, \textit{IMDB}, \textit{CivilComments}, and \textit{RAFT} use accuracy; \textit{NarrativeQA}, \textit{NaturalQuestions (closed-book)}, \textit{NaturalQuestions (closed-book)}, and \textit{QuAC} use F1 score; \textit{CNN/DailyMail} and \textit{XSUM} use ROUGE-2 score; \textit{MS MARCO (regular)} and \textit{MS MARCO} use RR@10 score and NDCG@10 score, respectively. For more details, please refer to \citet{helm}. In \ours, the evaluation score for \textit{HELM} is the average value of these 16 subtasks' scores.

\begin{figure*}[ht]
	\centering
        \begin{subfigure}{0.48\linewidth}
		\centering
		\includegraphics[width=1.0\linewidth]{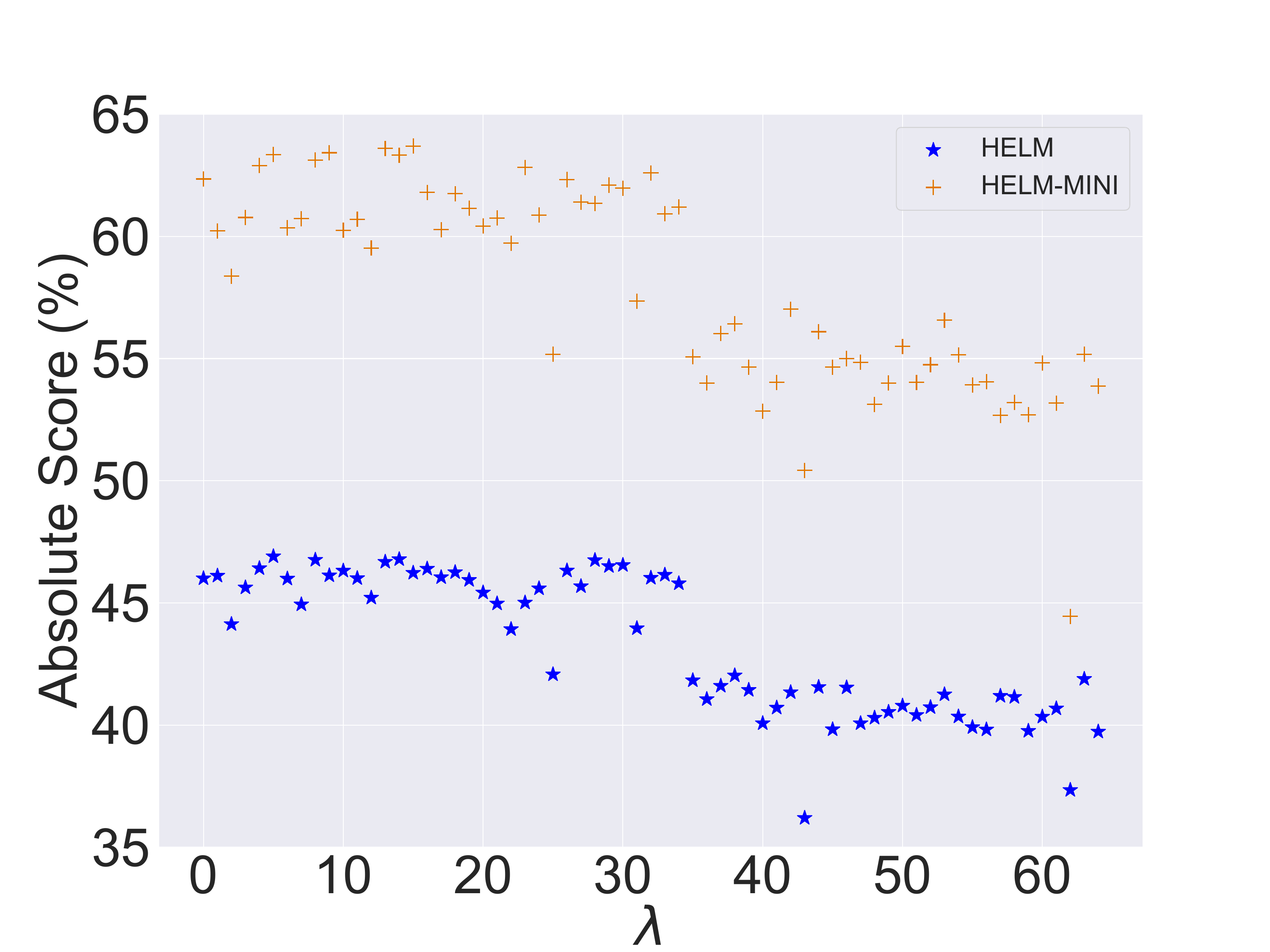}
		\caption{The average scores of the 16 subtasks in \textit{HELM} and the 4 subtasks in \textit{HELM-MINI} for the samples with different configurations, respectively.}
		\label{fig:helm_mini_value}
	\end{subfigure}
        \hspace{0.1in}
        \begin{subfigure}{0.48\linewidth}
		\centering
		\includegraphics[width=1.0\linewidth]{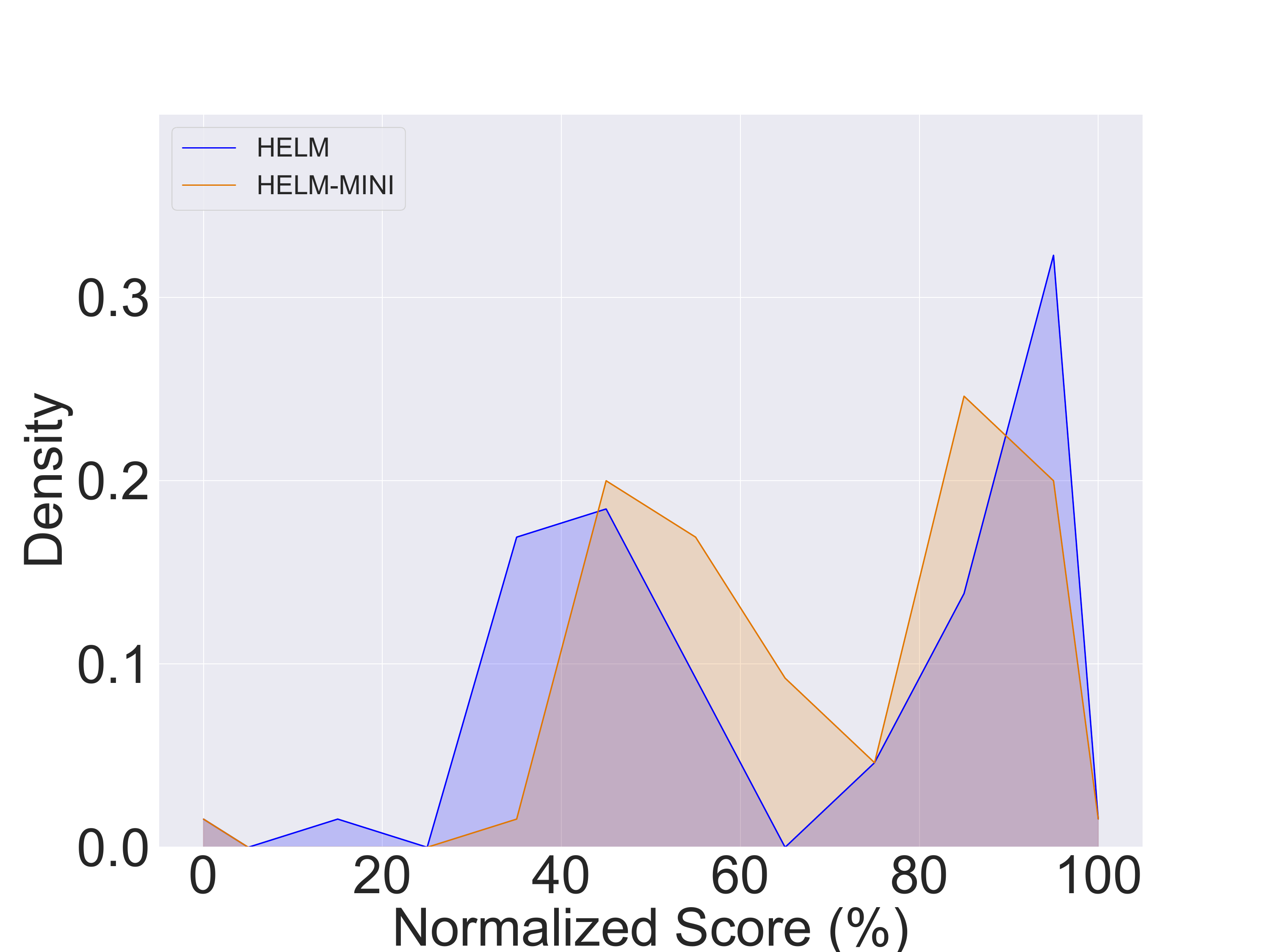}
		\caption{The density distribution of the normalized performance of \textit{HELM} and \textit{HELM-MINI} among the samples, respectively.}
		\label{fig:helm_mini_dis}
	\end{subfigure} 
	\caption{The performance comparison of \textit{HELM} and \textit{HELM-MINI}.}
	\label{fig:helm_mini}
\end{figure*}

However, evaluation of all these 16 subtasks in \textit{HELM} is very time-consuming. We notice that there is a trade-off between the evaluation's comprehensiveness and efficiency.
The more subtasks we use, the more accurate the evaluation scores will be, but also more time-consuming. 
Thus, we build \textit{HELM-MINI} with fewer subtasks to assess the LLMs' generic language capabilities.
We first randomly and uniformly sample serval samples of different combinations of configurations, including different PEFT algorithms and different hyperparameters.
After that, we pick 6 subtasks such that the evaluation score of these 6 subtasks is closest to the evaluation score of all 16 subtasks in the $L^2$ norm, which are \textit{MMLU}, \textit{NaturalQA (open-book)}, \textit{OpenbookQA}, \textit{MS MARCO (regular)}, \textit{XSUM} and \textit{IMDB}. 
Then we consider the time consumption and the requirements for the stability of the network connection during the evaluation, and drop another 2 subtasks: \textit{MS MARCO (regular)} and \textit{XSUM}. 
Finally, we have left 4 subtasks to form \textit{HELM-MINI}, including \textit{MMLU}, \textit{NaturalQA (open-book)}, \textit{OpenbookQA}, and \textit{IMDB}.
In Figure~\ref{fig:helm_mini}, we present the comparison between the evaluation scores of \textit{HELM} and \textit{HELM-MINI}. 
In Figure~\ref{fig:helm_mini_value}, we plot the samples of different combinations of configurations, where $\lambda$ indicates the index of each sample. 
It can be seen that though the exact evaluation scores of \textit{HELM} and \textit{HELM-MINI} are different, their normalized evaluation score relationships are essentially consistent. In Figure~\ref{fig:helm_mini_dis}, we show the distribution of these samples' normalized evaluation scores of \textit{HELM} and \textit{HELM-MINI}. Here, the normalized evaluation score is calculated by 
$\frac{x_i-x_{\min}}{x_{\max}-x_{\min}}$, where $x_{\max}$ and $x_{\min}$ are the maximum and minimum evaluation scores among all the samples of \textit{HELM} and \textit{HELM-MINI}, respectively. 

\noindent\textbf{Evaluation task for CoT capability.} We use \textit{GSM8K-test}~\citep{cobbe2021gsm8k}, which consists of 1k math problems from the test split in \textit{GSM8K} to evaluate the performance of LLMs on mathematic problem-solving capability. 
We also adopt 8-shot-CoT prompting~\citep{wei2022chain} in evaluation. The evaluation score is the accuracy of these 1k math problems. 

\subsection{Detailed hyperparameters and results}
\label{app:details}
We perform a grid search over the hyperparameter space to ensure that each algorithm achieves its optimal performance and to enable a fair comparison among different algorithms.
The search space of the learning rate for all adapters is $\{\num{1e-4}, \num{3e-4}, \num{5e-4}, \num{1e-3}, \num{3e-3}, \num{5e-3}\}$.

LoRA has three hyperparameters: the rank of the adaption matrices, the scaling coefficient, and the dropout probability for LoRA layers. 
The rank has a significant impact on resource consumption, so we fix it to 8, which is the default value in the algorithm. 
We grid search the other two parameters. 
The search spaces of the scaling coefficient and dropout probability for LoRA layers are $\{16, 32, 64, 128\}$ and $\{0.0, 0.1\}$, respectively.

For P-tuning, because of the source limitation, we fix the type of reparameterization to be ``MLP'' and only search the number of virtual tokens in $\{10, 20, 30\}$.

For prompt tuning, we find that initializing the prompt randomly will result in poor performances on these datasets.
Therefore, we set it to ``TEXT'' and initialize different prompts for different fine-tuning datasets. Specifically, 
for \textit{Fed\databar CodeAlpaca}, we use ``Program a function by following description.'' together with the number of virtual tokens equals 8;
For \textit{Fed\databar Dolly}, we use ``Assume you can understand and answer questions.'' together with the number of virtual tokens equals 9; 
For \textit{Fed\databar GSM8K-3}, we use ``Think step by step.'' together with the number of virtual tokens equals 6. 
The reason why we use the prompts above is that we think these prompts appropriately describe the fine-tuning tasks, respectively.
The number of virtual tokens we use is exactly the length of the corresponding prompt processed by the tokenizer.

In Table~\ref{tab:hpo_lora_codealpaca}-~\ref{tab:hpo_prompt_tuing_gsm8k}, we present experimental results when we grid search the hyperparameters for different PEFT algorithms.
In Table~\ref{tab:dia}, we show examples of creative writing generated by the fine-tuned LLaMA-7B with different PEFT algorithms. The fine-tuning dataset is \textit{Fed\databar Dolly}. It can be seen that the LLM, without fine-tuning in FL, lacks imagination and only describes the situation of a crocodile on the moon. The response of the model fine-tuned with LoRA is more accurate, which prefers the ocean to outer space and does write a haiku in the voice of a pirate. The model fine-tuned with P-tuning combines crocodiles and the moon, but the logic of the generated answers is not very coherent. The response generated by the model fine-tuned with prompt tuning is the shortest one, showing the savagery of the crocodile as a pirate. In summary, after federated fine-tuning, the accuracy and fluency of the LLM in answering questions are improved.

\begin{table*}[ht]
\renewcommand{\arraystretch}{1.5}
\centering
\caption{Evaluation scores (\%): federated fine-tuning LLaMA-7B with LoRA on \textit{Fed\databar CodeAlpaca}. In this table, the dropout probability is 0.0.}
\label{tab:hpo_lora_codealpaca}
\begin{tabular}{cccccccc}
\toprule
\multicolumn{1}{l}{Scaling coef.} &\multicolumn{1}{l}{Seed}& \multicolumn{6}{c}{LR}  \\  \cline{3-8}
 &  & 0.0001 & 0.0003 & 0.0005 & 0.001 & 0.003 & 0.005 \\ \hline
\multirow{3}{*}{16}  &  0 & 13.29 & 11.34 & 10.37 & 9.39 & 11.34 & 10.73 \\ 
  &  1 & 13.17 & 11.71 & 10.49 & 10.73 & 11.71 & 11.46 \\ 
  & 2 & 13.41 & 11.22 & 11.10 & 10.85 & 12.68 & 11.83 \\ \hline
\multirow{3}{*}{32} &  0   & 12.56 & 10.24 & 9.27 & 10.61 & 10.24 & 2.07 \\
   & 1 & 12.07 & 11.22 & 10.98 & 10.49 & 8.54 & 9.88 \\ 
   & 2 & 12.44 & 11.34 & 11.22 & 11.34 & 10.73 & 10.00 \\ \hline
\multirow{3}{*}{64}  &  0 & 10.00 & 9.63 & 11.10 & 11.10 & 10.24 & 11.22 \\
 & 1 & 10.73 & 10.37 & 11.22 & 11.10 & 10.98 & 11.71 \\ 
   & 2 & 11.10 & 12.56 & 12.68 & 12.44 & 9.76 & 12.68 \\ \hline
\multirow{3}{*}{128}   & 0 & 9.88 & 10.98 & 11.83 & 11.83 & 10.85 & 10.73 \\ 
  & 1 & 10.24 & 12.07 & 11.83 & 11.95 & 0.00 & 10.00 \\ 
 & 2 & 10.98 & 12.68 & 11.95 & 9.15 & 6.83 & 5.85 \\ \bottomrule
\end{tabular}
\end{table*}

\begin{table*}[ht]
\renewcommand{\arraystretch}{1.5}
\centering
\caption{Evaluation scores (\%): federated fine-tuning LLaMA-7B with P-tuning on \textit{Fed\databar CodeAlpaca}.}
\label{tab:hpo_p_tuning_codealpaca}
\begin{tabular}{cccccccc}
\toprule
\multicolumn{1}{c}{\#Virtual tokens} & \multicolumn{1}{c}{Seed}  & \multicolumn{6}{c}{LR}  \\  \cline{3-8} 
& & 0.0001 & 0.0003 & 0.0005 &  0.001 & 0.003 & 0.005 \\ \hline
\multirow{3}{*}{10} & 0 & 8.90 & 6.83 & 8.78 & 10.24 & 10.61 & 6.83 \\
 & 1 & 11.10 & 9.15 & 10.00 & 9.51 & 1.46 & 6.95 \\
 & 2 & 5.37 & 9.39 & 5.85 & 4.14 & 5.37 & 1.83 \\ \hline
\multirow{3}{*}{20} & 0 & 5.73 & 11.22 & 9.39 & 13.05 & 9.51 & 4.51 \\
 & 1 & 8.78 & 10.85 & 10.00 & 9.39 & 9.02 & 3.05 \\
 & 2 & 7.93 & 3.17 & 8.78 & 2.93 & 10.61 & 3.66 \\ \hline
\multirow{3}{*}{30} & 0 & 9.02 & 10.24 & 10.00 & 10.98 & 9.14 & 1.95 \\ 
 & 1 & 12.32 & 5.49 & 5.37 & 10.24 & 2.68 & 2.80 \\
 & 2 & 7.07 & 9.02 & 9.63 & 5.24 & 10.61 & 0.24 \\ \bottomrule
\end{tabular}
\end{table*}

\begin{table*}[ht]
\renewcommand{\arraystretch}{1.5}
\centering
\caption{Evaluation scores (\%): federated fine-tuning LLaMA-7B with prompt tuning on \textit{Fed\databar CodeAlpaca}. In this table, the initialized prompt is ``Program a function by following description.''. The number of virtual tokens is 8.}
\label{tab:hpo_prompt_tuing_codealpaca}
\begin{tabular}{ccccccc}
\toprule
\multicolumn{1}{c}{Seed} & \multicolumn{6}{c}{LR} \\ \cline{2-7}
   & 0.0001 & 0.0003 & 0.0005 & 0.001 & 0.003 & 0.005 \\ \hline
  0 & 10.16 & 11.07 & 10.69 & 9.63 & 12.28 & 14.03 \\
  1 & 7.96 & 9.40 & 10.61 & 11.90 & 11.60 & 9.63 \\
  2 & 8.64 & 9.40 & 10.24 & 10.39 & 11.37 & 11.22 \\ \bottomrule
\end{tabular}
\end{table*}

\begin{table*}[ht]
\renewcommand{\arraystretch}{1.5}
\centering
\caption{Evaluation scores (\%): federated fine-tuning LLaMA-7B with LoRA on \textit{Fed\databar Dolly}.}
\label{tab:hpo_lora_dolly}
\scalebox{0.93}{
\begin{tabular}{cccccccccc}
\toprule
Scaling coef. & \multicolumn{3}{c}{32} & \multicolumn{3}{c}{64} & \multicolumn{3}{c}{128} \\
Dropout prob. & \multicolumn{3}{c}{0.0} & \multicolumn{3}{c}{0.1} & \multicolumn{3}{c}{0.0} \\
LR & \multicolumn{3}{c}{0.005} & \multicolumn{3}{c}{0.003} & \multicolumn{3}{c}{0.0003} \\ \cline{2-10}
Seed & 0 & 1 & 2 & 0 & 1 & 2 & 0 & 1 & 2 \\ \hline
\textit{MMLU} & 36.30 & 35.50 & 36.80 & 36.50 & 37.20 & 37.40 & 33.20 & 35.20 & 33.80 \\
\textit{BoolQ} & 80.00 & 77.00 & 78.00 & 82.00 & 81.00 & 74.00 & 78.00 & 79.00 & 80.00 \\
\textit{NarrativeQA} & 55.60 & 49.80 & 55.00 & 53.80 & 51.60 & 53.20 & 51.00 & 51.60 & 50.90 \\
\textit{NaturalQuestions(closed)} & 21.20 & 22.00 & 20.40 & 21.40 & 28.00 & 20.60 & 23.70 & 22.20 & 23.80 \\
\textit{NaturalQuestions(open)} & 71.30 & 66.90 & 61.60 & 70.90 & 67.10 & 69.40 & 64.00 & 60.90 & 63.20 \\
\textit{QuAC} & 32.30 & 31.20 & 31.20 & 32.40 & 28.40 & 28.50 & 32.10 & 34.30 & 32.60 \\
\textit{HellaSwag} & 81.00 & 82.00 & 80.00 & 82.00 & 79.00 & 79.00 & 81.00 & 81.00 & 81.00 \\
\textit{OpenbookQA} & 50.00 & 53.00 & 46.00 & 52.00 & 55.00 & 51.00 & 52.00 & 49.00 & 52.00 \\
\textit{TruthfulQA} & 18.00 & 30.00 & 30.00 & 24.00 & 24.00 & 23.00 & 23.00 & 30.00 & 25.00 \\
\textit{MS MARCO (regular)} & 18.60 & 14.20 & 17.40 & 18.20 & 15.50 & 22.90 & 17.20 & 15.90 & 15.00 \\
\textit{MS MARCO (TREC)} & 40.60 & 42.70 & 42.50 & 38.50 & 41.20 & 42.70 & 41.40 & 41.90 & 39.70 \\
\textit{CNN/DailyMail} & 15.10 & 12.30 & 14.00 & 13.60 & 14.40 & 13.00 & 12.00 & 12.10 & 11.80 \\
\textit{XSUM} & 9.50 & 10.10 & 10.30 & 13.70 & 10.90 & 10.80 & 8.80 & 9.50 & 11.80 \\
\textit{IMDB} & 94.00 & 98.00 & 97.00 & 95.00 & 94.00 & 97.00 & 98.00 & 96.00 & 98.00 \\
\textit{CivilComments} & 59.50 & 66.00 & 56.90 & 57.00 & 59.70 & 58.30 & 62.20 & 57.30 & 61.70 \\
\textit{RAFT} & 59.70 & 59.80 & 58.80 & 55.90 & 61.60 & 58.90 & 64.80 & 60.90 & 59.80 \\ \hline
\textbf{Average} & 46.42 & 46.91 & 45.99 & 46.68 & 46.79 & 46.23 & 46.40 & 46.05 & 46.26\\ \bottomrule
\end{tabular}}
\end{table*}

\begin{table*}[ht]
\renewcommand{\arraystretch}{1.5}
\centering
\caption{Evaluation scores (\%): federated fine-tuning LLaMA-7B with P-tuning on \textit{Fed\databar Dolly}.}
\label{tab:hpo_t_tuning_dolly}
\scalebox{0.93}{
\begin{tabular}{cccccccccc}
\toprule
\#Virtual tokens & \multicolumn{3}{c}{20} & \multicolumn{3}{c}{20} & \multicolumn{3}{c}{30} \\
LR & \multicolumn{3}{c}{0.0003} & \multicolumn{3}{c}{0.0005} & \multicolumn{3}{c}{0.0005} \\ \cline{2-10}
Seed & \multicolumn{1}{c}{0} & \multicolumn{1}{c}{1} & \multicolumn{1}{c}{2} & \multicolumn{1}{c}{0} & \multicolumn{1}{c}{1} & \multicolumn{1}{c}{2} & 0 & 1 & 2 \\ \hline
\textit{MMLU} & 35.20 & 34.60 & 35.20 & 35.30 & 35.10 & 35.10 & 34.30 & 35.40 & 35.10 \\
\textit{BoolQ} & 79.00 & 78.00 & 78.00 & 77.00 & 80.00 & 76.00 & 82.00 & 78.00 & 79.00 \\
\textit{NarrativeQA} & 50.40 & 54.60 & 52.60 & 52.30 & 52.10 & 49.70 & 52.20 & 52.30 & 52.50 \\
\textit{NaturalQuestions(closed)} & 26.20 & 25.50 & 25.00 & 25.40 & 27.60 & 25.50 & 25.60 & 24.90 & 27.10 \\
\textit{NaturalQuestions(open)}  & 66.80 & 66.10 & 66.60 & 67.10 & 66.20 & 64.00 & 66.80 & 65.90 & 65.60 \\
\textit{QuAC} & 28.80 & 30.00 & 29.50 & 30.10 & 27.70 & 29.80 & 31.10 & 30.50 & 30.20 \\
\textit{HellaSwag} & 22.00 & 22.00 & 20.00 & 25.00 & 23.00 & 21.00 & 17.00 & 20.00 & 19.00 \\
\textit{OpenbookQA} & 23.30 & 23.30 & 27.30 & 30.30 & 23.30 & 18.30 & 28.30 & 22.30 & 25.30 \\
\textit{TruthfulQA} & 30.00 & 32.00 & 33.00 & 29.00 & 28.00 & 35.00 & 29.00 & 26.00 & 28.00 \\
\textit{MS MARCO (regular)} & 18.10 & 17.60 & 16.00 & 15.70 & 15.50 & 13.90 & 20.50 & 12.80 & 16.40 \\
\textit{MS MARCO (TREC)}  & 45.00 & 39.50 & 42.60 & 43.50 & 42.40 & 36.40 & 41.00 & 30.00 & 44.00 \\
\textit{CNN/DailyMail} & 12.90 & 11.70 & 13.70 & 12.40 & 13.40 & 12.40 & 13.50 & 13.50 & 12.30 \\
\textit{XSUM} & 11.30 & 10.00 & 10.50 & 11.30 & 11.60 & 11.10 & 10.60 & 9.50 & 10.70 \\
\textit{IMDB} & 95.00 & 92.00 & 95.00 & 93.00 & 94.00 & 94.00 & 95.00 & 95.00 & 94.00 \\
\textit{CivilComments} & 63.70 & 59.40 & 59.60 & 61.90 & 62.20 & 57.40 & 56.60 & 59.70 & 63.10 \\
\textit{RAFT} & 61.60 & 60.70 & 61.10 & 63.20 & 60.90 & 61.60 & 61.40 & 61.40 & 62.30 \\ \hline
\textbf{Average} & 41.83 & 41.06 & 41.61 & 42.03 & 41.44 & 40.08 & 41.56 & 39.83 & 41.54 \\ \bottomrule
\end{tabular}}
\end{table*}

\begin{table*}[ht]
\renewcommand{\arraystretch}{1.5}
\centering
\caption{Evaluation scores (\%): federated fine-tuning LLaMA-7B with prompt tuning on \textit{Fed\databar Dolly}. In this table, the initialized prompt is “Assume you can understand and answer questions.”. The number of virtual tokens is 9.}
\label{tab:hpo_prompt_tuning_dolly}
\scalebox{0.93}{
\begin{tabular}{cccccccccc}
\toprule
LR & \multicolumn{3}{c}{0.0003} & \multicolumn{3}{c}{0.0005} & \multicolumn{3}{c}{0.001} \\ \cline{2-10}
Seed & 0 & 1 & 2 & 0 & 1 & 2 & 0 & 1 & 2 \\ \hline
\textit{MMLU} & 34.10 & 34.50 & 34.00 & 35.10 & 35.50 & 34.60 & 36.00 & 34.40 & 34.20 \\
\textit{BoolQ} & 79.00 & 77.00 & 76.00 & 77.00 & 78.00 & 74.00 & 77.00 & 77.00 & 76.00 \\
\textit{NarrativeQA} & 50.10 & 49.20 & 50.10 & 50.50 & 51.50 & 52.10 & 50.10 & 52.70 & 50.40 \\
\textit{NaturalQuestions(closed)}  & 24.70 & 24.40 & 24.90 & 23.50 & 25.80 & 24.30 & 23.80 & 26.30 & 25.80 \\
\textit{NaturalQuestions(open)} & 66.60 & 64.60 & 66.00 & 64.90 & 62.80 & 65.80 & 61.90 & 63.00 & 64.60 \\
\textit{QuAC} & 32.00 & 28.50 & 28.40 & 29.50 & 29.00 & 29.60 & 30.20 & 30.10 & 28.60 \\
\textit{HellaSwag} & 19.00 & 19.00 & 20.00 & 21.00 & 19.00 & 20.00 & 21.00 & 20.00 & 19.00 \\
\textit{OpenbookQA} & 27.30 & 20.00 & 21.00 & 32.30 & 27.30 & 20.30 & 27.30 & 19.30 & 19.00 \\
\textit{TruthfulQA} & 32.00 & 32.00 & 34.00 & 32.00 & 31.00 & 33.00 & 29.00 & 36.00 & 33.00 \\
\textit{MS MARCO (regular)} & 13.60 & 14.50 & 13.70 & 14.10 & 14.20 & 14.10 & 12.70 & 17.70 & 21.80 \\
\textit{MS MARCO (TREC)} & 35.90 & 41.20 & 39.00 & 39.70 & 39.60 & 35.30 & 39.80 & 43.20 & 43.70 \\
\textit{CNN/DailyMail} & 12.30 & 13.00 & 13.70 & 13.60 & 14.20 & 13.90 & 13.30 & 14.10 & 14.40 \\
\textit{XSUM} & 10.30 & 12.10 & 10.40 & 11.70 & 11.20 & 10.70 & 10.70 & 11.20 & 11.10 \\
\textit{IMDB} & 94.00 & 97.00 & 98.00 & 94.00 & 95.00 & 95.00 & 91.00 & 94.00 & 95.00 \\
\textit{CivilComments} & 61.30 & 59.10 & 62.30 & 62.60 & 54.70 & 56.60 & 58.50 & 59.90 & 60.00 \\
\textit{RAFT} & 60.50 & 60.50 & 60.20 & 58.60 & 56.80 & 59.30 & 54.80 & 60.20 & 61.80 \\ \hline
\textbf{Average} & 40.79 & 40.41 & 40.73 & 41.26 & 40.35 & 39.91 & 39.82 & 41.19 & 41.15 \\ \bottomrule
\end{tabular}}
\end{table*}

\begin{table*}[ht]
\renewcommand{\arraystretch}{1.5}
\centering
\caption{Evaluation scores (\%): federated fine-tuning LLaMA-7B with LoRA on \textit{Fed\databar GSM8K-3}.}
\label{tab:hpo_lora_gsm8k}
\begin{tabular}{cccccccc}
\toprule
\multicolumn{1}{l}{Scaling coef.} &\multicolumn{1}{l}{Seed}& \multicolumn{6}{c}{LR}  \\  \cline{3-8}
 &  & 0.0001 & 0.0003 & 0.0005 & 0.001 & 0.003 & 0.005 \\ \hline
\multirow{3}{*}{16} & 0 & 8.79 & 6.97 & 8.87 & 8.26 & 12.13 & 11.22 \\
 & 1 & 8.34 & 8.19 & 8.95 & 8.95 & 9.55 & 13.04 \\
 & 2 & 9.17 & 6.37 & 9.63 & 9.33 & 11.83 & 13.42 \\ \hline
\multirow{3}{*}{32} & 0 & 6.82 & 7.73 & 9.25 & 12.21 & 10.99 & 4.09 \\
 & 1 & 7.81 & 8.04 & 8.79 & 10.16 & 13.19 & 0.00 \\
 & 2 & 9.10 & 8.79 & 8.95 & 11.9 & 12.89 & 13.87 \\ \hline
\multirow{3}{*}{64} & 0 & 8.95 & 10.61 & 11.30 & 12.13 & 13.42 & 0.00 \\
 & 1 & 8.26 & 9.40 & 11.30 & 10.99 & 0.00 & 0.30 \\
 & 2 & 9.93 & 10.46 & 12.28 & 12.43 & 12.28 & 13.42 \\ \hline
\multirow{3}{*}{128} & 0 & 9.93 & 11.14 & 12.96 & 3.49 & 15.31 & 15.47 \\
 & 1 & 10.54 & 9.40 & 13.12 & 10.99 & 15.09 & 10.92 \\
 & 2 & 9.25 & 11.83 & 11.60 & 12.96 & 12.28 & 11.60 \\ \bottomrule
\end{tabular}
\end{table*}

\begin{table*}[ht]
\renewcommand{\arraystretch}{1.5}
\centering
\caption{Evaluation scores (\%): federated fine-tuning LLaMA-7B with P-tuning on \textit{Fed\databar GSM8K-3}.}
\label{tab:hpo_p_tuning_gsm8k}
\begin{tabular}{cccccccc}
\toprule
\multicolumn{1}{l}{\#Virtual tokens} & \multicolumn{1}{c}{Seed} & \multicolumn{6}{c}{LR}  \\  \cline{3-8} 
& & 0.0001 & 0.0003 & 0.0005 & 0.001 & 0.003 & 0.005 \\ \hline
\multirow{3}{*}{10} & 0 & 10.16 & 11.07 & 10.69 & 9.63 & 12.28 & 14.03 \\
 & 1 & 7.96 & 9.40 & 10.61 & 11.90 & 11.60 & 9.63 \\
 & 2 & 8.64 & 9.40 & 10.24 & 10.39 & 11.37 & 11.22 \\ \hline
\multirow{3}{*}{20} & 0 & 9.40 & 11.22 & 12.74 & 11.75 & 9.40 & 10.24 \\
 & 1 & 8.26 & 9.17 & 10.08 & 11.60 & 12.36 & 8.72 \\
 & 2 & 10.39 & 10.54 & 10.77 & 10.01 & 10.77 & 11.68 \\ \hline
\multirow{3}{*}{30} & 0 & 10.08 & 11.14 & 12.05 & 11.07 & 10.69 & 8.19 \\
 & 1 & 6.60 & 9.63 & 9.33 & 9.25 & 10.16 & 10.84 \\
 & 2 & 8.87 & 9.93 & 10.77 & 10.46 & 10.39 & 11.22 \\ \bottomrule
\end{tabular}
\end{table*}

\begin{table*}[ht]
\renewcommand{\arraystretch}{1.5}
\centering
\caption{Evaluation scores (\%): federated fine-tuning LLaMA-7B with prompt tuning on \textit{Fed\databar GSM8K-3}. In this table, the initialized prompt is “Think step by step.”. The number of virtual tokens is 6.}
\label{tab:hpo_prompt_tuing_gsm8k}
\begin{tabular}{ccccccc}
\toprule
  \multicolumn{1}{c}{Seed} & \multicolumn{6}{c}{LR} \\ \cline{2-7}
  & 0.0001 & 0.0003 & 0.0005 & 0.001 & 0.003 & 0.005 \\ \hline
 0 & 8.64 & 10.01 & 9.33 & 10.08 & 9.70 & 9.93 \\
 1 & 8.19 & 9.33 & 9.78 & 9.10 & 7.28 & 9.10 \\
 2 & 8.79 & 8.87 & 8.72 & 9.40 & 9.02 & 10.54 \\ \bottomrule
\end{tabular}
\end{table*}

\begin{table*}[ht]
\renewcommand{\arraystretch}{1.5}
\centering
\begin{tabular}{ll}
\toprule
User: & Write a haiku about crocodiles in outer space in the voice of a pirate. \\ \hline
\multirow{2}{*}{System0:} & I’m a croc in space, And I don’t know what to do. I’ve never been here before, And \\
 & it’s all so new to me. But I know I have to find a way To get back home to you. \\ \cline{2-2} 
\multirow{4}{*}{System1:} & Crocodiles, in space? I’d rather be on the sea. Pirate’s life for me! -Arrrrrrr! Hoist \\
 & the Jolly Roger, matey. We’re off to plunder the seven seas! Ahoy, me hearties! \\
 & Let the pillaging begin! -Aye, aye Cap’n! - Soon we’ll be rich beyond our wildest \\
 & dreams, and we won’t have to work a day in our lives. \\ \cline{2-2} 
\multirow{2}{*}{System2:} & Crocodiles on the moon, Ahoy me hearties! We’ll be pirates no more! The crocs \\
 & have won! Mwahahaha. Aye, matey! Hoist the Jolly Roger! \\ \cline{2-2} 
System3: & Crocodiles on the moon, I’m going to eat you all! Ahoy, matey! \\ \bottomrule
\end{tabular}
\caption{Comparison of creative writing capability among different fine-tuning algorithms for LLaMA-7B in FL. System0 is the baseline model without fine-tuning, System1 is fine-tuned with LoRA, System2 is fine-tuned with P-tuning, and System3 is fine-tuned with prompt tuning.}
\label{tab:dia}
\end{table*}

\end{document}